%% file: main.tex
\documentclass[10pt,twocolumn,letterpaper]{article}

\usepackage[pagenumbers]{cvpr} 

\input{preamble}

\definecolor{cvprblue}{rgb}{0.21,0.49,0.74}
\usepackage[pagebackref,breaklinks,colorlinks,citecolor=cvprblue]{hyperref}
\usepackage[capitalize]{cleveref}

\input{def.tex}
\def\paperID{1551} 
\def\confName{CVPR}
\def\confYear{2024}

\title{Multi-criteria Token Fusion with One-step-ahead Attention \\for Efficient Vision Transformers}

\newcommand*\samethanks[1][\value{footnote}]{\footnotemark[#1]}
\author{\textbf{Sanghyeok Lee}\thanks{Equal contribution.} \hspace{0.4cm} \textbf{Joonmyung Choi}\samethanks \hspace{0.4cm}  \textbf{Hyunwoo J. Kim}\thanks{Corresponding author.} \vspace{0.4cm}
\\
Department of Computer Science and Engineering, Korea University\\
{\tt\small \{\href{mailto:cat0626@korea.ac.kr}{cat0626}, \href{mailto:pizard@korea.ac.kr}{pizard}, \href{mailto:hyunwoojkim@korea.ac.kr}{hyunwoojkim}\}@korea.ac.kr}
}

\begin{document}
\maketitle
\input{section/abstract}
\input{section/introduction}

\input{section/relatedworks}
\input{section/method}

\input{section/experiments}

\input{section/conclusion}
{
    \small
    \bibliographystyle{ieeenat_fullname}
    \bibliography{main}
}

\onecolumn
\setcounter{table}{0}
\renewcommand{\thetable}{\Alph{table}}
\setcounter{figure}{0}
\renewcommand{\thefigure}{\Alph{figure}}

\appendix
\clearpage
\section{Implementation details}
For a comparison with previous works, we first evaluate MCTF with DeiT~\cite{touvron2021training} on ImageNet-1K~\cite{deng2009imagenet}. Following~\cite{rao2021dynamicvit, kong2022spvit,yin2022vit}, we finetune the model with the pre-trained weights for 30 epochs with the batch size of 1,024 under 8 RTX3090 GPUs. 
We opt for the least epochs among previous works (\eg, 30 for DynamicViT~\cite{rao2021dynamicvit}, 60 for SPViT~\cite{kong2022spvit}, 100 for A-ViT~\cite{yin2022vit}).
For finetuning, the learning rate is initially set to 3e-5 and decreases to 1e-6 by the cosine annealing~\cite{loshchilov2016sgdr} with a cooldown of 10 epochs.
Also, we finetune the T2T-ViT~\cite{yuan2021tokens} and LV-ViT~\cite{jiang2021all} with the initial learning rate of 5e-6, 1e-5 decreasing to 5e-7, 2e-6 for 30 epochs followed by 10 cooldown epochs, respectively.
We do not use mixup-based augmentation~\cite{DBLP:conf/iclr/ZhangCDL18,yun2019cutmix} to prevent the corrupted representation in fused tokens caused by the token fusion between different samples.
Since we already track the size of the tokens, we also adopt proportional attention of ToMe~\cite{bolya2022token} which simply update the attention scores with the size of the tokens $\textbf{s}$ as $\mathbf{A} = \text{softmax} \left( \frac{\mathbf{QK}^\top}{\sqrt{C}} + \log \mathbf{s} \right)$.
Regarding hyper-parameters for MCTF, we use $[\tau^\text{info},\tau^\text{sim},\tau^\text{size}] = [1, 1/20, 1/40]$ for the temperature parameters. 
And, We opt $\lambda = 1$ for DeiT-T and T2T-ViT, and $\lambda=3$ for DeiT-S and LV-ViT for the coefficient of consistency loss. Similar to UDA~\cite{xie2020unsupervised}, the consistency loss is calculated only with the sample that has a confidence score higher than $\beta=0.4$.
We also set the safeguard for excessive fusion by maintaining at least 10 tokens.
For measuring the efficiency, we use \texttt{fvcore} and report the FLOPs of the model. 

\section{Analyses on MCTF}
\subsection{Sensitivity analysis on hyper-parameters of MCTF}
To analyze the sensitivity of the hyper-parameters in MCTF, we compare the accuracy according to the temperature parameter $\tau$ in~\Cref{tab:sup5}. 
While evaluating each parameter, other hyper-parameters are set to default values mentioned in the implementation details. 
We run the experiments with DeiT-S equipped with MCTF ($r=16$). 
The default settings for each hyper-parameter are highlighted.
\input{tables/table_sup5}

\subsection{Loss of information}
In this subsection, we measure the loss of information to validate the efficacy of MCTF. For this, we consider the cosine similarity between the class tokens with and without MCTF ($r=16$) as a metric to measure the loss of information, which indicates the changes in the class tokens.
In other words, if the similarity between class tokens is low, we infer that the fused tokens significantly affect the class token's representation while losing the information of original contents. 
The differences between the class tokens at each block are reported in~\Cref{tab:sup4}.
As shown in the table, at the early stage of the Transformer (\eg, [1-6]-th block), there is no big gap among the diverse criteria. 
However, as the number of fused tokens increases through consecutive blocks, there are substantial changes in the class tokens.
Specifically, when we consider a single criterion, similarity is the best option for mitigating the loss of information compared to informativeness and size.
Then, adopting the dual criterion composed of similarity and informativeness, we further lessen the changes between the class tokens showing the high similarity even in the rear block (\eg, [7-12]-th block). 
At last, MCTF with all three criteria shows better similarity than dual-criteria. 
We believe that this minimization of information loss by adopting multi-criteria leads to consistent improvements compared to other single and dual criteria in image classification.
\input{tables/table_sup4}
\clearpage
\subsection{Qualitative comparison for one-step-ahead attention}
In MCTF, the attention map $\hat{\mathbf{A}}^{l+1}$ of the fused tokens $\hat{\mathbf{X}}^{l}$ is approximated by aggregating the one-step-ahead attention $\mathbf{A}^{l+1}$, which is the attention before token fusion.
In \Cref{sec:4.2}, we shows that this approximation brings substantial speed improvements without any performance degradation by avoiding the re-computation of self-attention. 
In parallel, we here provide a qualitative comparison to show the soundness of our approaches. The visualization of the attention map in the [3,6,9,12]-th layer is provided in \Cref{fig:onestepahead_qa}.
\input{Figures/onestepahead_qa}

\clearpage
\section{Detailed results}
\label{sec:sup1}
In this section, we provide more detailed results of MCTF with the Vision Transformers in ImageNet-1K~\cite{deng2009imagenet}. 
\subsection{Full results with DeiT~\cite{touvron2021training}.}
As the settings in the ablations studies, we first finetune the model with $r = 16$ for the number of reduced tokens per layer and report the flops and accuracies with varying $r$. We highlight the row used for finetuning. Also, we present the detailed results of MCTF without any additional training. Full results with and without finetuning are summarized in~\Cref{tab:sup1} and~\Cref{tab:sup2}, respectively.
\input{tables/table_sup1}
\input{tables/table_sup2}
\clearpage
\subsection{Full results with T2T-ViT~\cite{yuan2021tokens} and LV-ViT~\cite{jiang2021all}.}
We also present the full results with T2T-ViT and LV-ViT in \Cref{tab:sup3}. Note that, similar to DeiT-S, we report the FLOPs and accuracies in varying reduction ratios with the model finetuned with a specific reduction ratio, which is used for reporting the results in~\Cref{tab:main2}. We also highlight this reduction ratio in the table. It is worth noting that, although each model is finetuned with the specific $r$, MCTF shows promising performance within the range from 1 to $r$.
\input{tables/table_sup3}

\end{document}



\section*{Supplement of Multi-criteria Token Fusion for Efficient Vision Transformers}
In this supplement, we delineate implementation details in~\Cref{sec:a}, and provide more experiments in~\Cref{sec:b} and analyses in~\Cref{sec:c}.
\input{supple/implementation}
\input{supple/experiments}

\input{supple/analyses}

\newpage
\bibliographystyle{unsrt} 
\bibliography{main/reference}

%% file: preamble.tex
\usepackage[dvipsnames]{xcolor}
\newcommand{\red}[1]{{\color{red}#1}}

\usepackage{url}

\usepackage{booktabs}       
\usepackage{amsfonts}       
\usepackage{nicefrac}       
\usepackage{microtype}      
\usepackage{colortbl}         

\usepackage{graphicx}
\usepackage{multirow}
\usepackage{amsmath}
\usepackage{subcaption}
\usepackage{graphicx}
\usepackage{algorithm}
\usepackage{algpseudocode}
\DeclareMathOperator*{\argmax}{arg\,max}

\newcommand{\green}{\textcolor[rgb]{0.1,0.9,0.1}}
\newcommand{\gray}{\textcolor[rgb]{0.5,0.5,0.5}}
\newcommand{\blue}{\textcolor[rgb]{0,0,1}}

\definecolor{Gray}{rgb}{0.8,0.8,0.8} 
\definecolor{LightBlue}{rgb}{0.8,0.9,1} 
\definecolor{LightOrange}{rgb}{1,0.9,0.8} 
\definecolor{Cyan}{rgb}{0,1,1}

%% file: def.tex
\newcommand{\xclsr}{\mathbf{x}^\text{cls}_r}

%% file: section/abstract.tex
\begin{abstract}
Vision Transformer (ViT) has emerged as a prominent backbone for computer vision. 
For more efficient ViTs, recent works lessen the quadratic cost of the self-attention layer by pruning or fusing the redundant tokens. 
However, these works faced the speed-accuracy trade-off caused by the loss of information. 
Here, we argue that token fusion needs to consider diverse relations between tokens to minimize information loss. 
In this paper, we propose a Multi-criteria Token Fusion (MCTF), that gradually fuses the tokens based on multi-criteria (\ie, similarity, informativeness, and size of fused tokens). 
Further, we utilize the one-step-ahead attention, which is the improved approach to capture the informativeness of the tokens.
By training the model equipped with MCTF using a token reduction consistency, we achieve the best speed-accuracy trade-off in the image classification (ImageNet1K). 
Experimental results prove that MCTF consistently surpasses the previous reduction methods with and without training. 
Specifically, DeiT-T and DeiT-S with MCTF reduce FLOPs by about 44\% while improving the performance (+0.5\%, and +0.3\%) over the base model, respectively. 
We also demonstrate the applicability of MCTF in various Vision Transformers (\eg, T2T-ViT, LV-ViT), achieving at least 31\% speedup without performance degradation.
Code is available at \url{https://github.com/mlvlab/MCTF}.

\end{abstract}

%% file: section/introduction.tex
\input{Figures/Figure1}

\section{Introduction}
\label{sec:1}
Vision Transformer~\cite{dosovitskiy2020image} (ViT) has been proposed to tackle the vision tasks with self-attention, originally developed for natural language processing tasks.
With the advent of ViT, Transformers are the prevalent architectures for a wide range of vision tasks, \textit{e.g.}, classification~\cite{dosovitskiy2020image, touvron2021training, touvron2021going, liu2021swin, wang2021pyramid}, object detection~\cite{wang2021pyramid, carion2020end, zhu2020deformable}, segmentation~\cite{liu2021swin, xie2021segformer,strudel2021segmenter}, etc.
ViTs, built only with self-attention and MLP, provide great flexibility and impressive performance compared to conventional approaches, \textit{e.g.}, convolutional neural networks (CNNs). 
However, despite these advantages, the quadratic computational complexity of self-attention with respect to the number of tokens is the major bottleneck for Transformers. 
This limitation becomes more substantial with the growing interest in large-scale foundation models such as CLIP~\cite{radford2021learning}. 
To this end, several works~\cite{kitaev2020reformer, wang2020linformer, xiong2021nystromformer, Beltagy2020Longformer} have proposed efficient self-attention mechanisms
including local self-attention within predefined windows~\cite{liu2021swin,chu2021twins,arar2022learned}.\\
\noindent More recently, there has been increasing interest in token-reduction methods for optimizing ViTs without altering their architecture.
Earlier works~\cite{meng2022adavit,pan2021ia,fayyaz2022adaptive, yin2022vit, rao2021dynamicvit} primarily focused on pruning the uninformative tokens to reduce the number of tokens. 
Another line of works~\cite{kong2022spvit,liang2022not,marin2023token,bolya2022token,long2023beyond} attempted to fuse the tokens instead of discarding them
to minimize the information loss. 
However, performance degradation is still commonly observed in most token fusion methods. 
We notice that the token fusion methods usually consider only one criterion, such as the similarity or informativeness of tokens, leading to suboptimal token matching. 
For instance, similarity-based token fusion is prone to combine the foreground tokens, whereas informativeness-based fusion 
often merges substantially dissimilar tokens, resulting in collapsed representations. 
Furthermore, if too many tokens are fused into one token, then information loss is inevitable.\\
\noindent To address the problems, we introduce \textbf{M}ulti-\textbf{C}riteria \textbf{T}oken \textbf{F}usion (\textbf{MCTF}) that optimizes 
vision transformers by fusing tokens based on multi-criteria.
Unlike previous works that consider a single criterion for token fusion, MCTF measures the relationship between the tokens with multi-criteria as follows; (1) similarity to fuse the redundant tokens, (2) informativeness to reduce the uninformative tokens, (3) the size of the tokens to prevent the large-sized tokens that boost the loss of information. 
Also, to tackle the inconsistency between attention maps of consecutive layers, we adopt \emph{one-step-ahead attention}, which explicitly estimates the informativeness of the tokens in the next layer. 
Finally, by introducing a \emph{token reduction consistency} for finetuning the model, we achieve superior performance to the existing works as in~\Cref{fig:comparison}. 
Surprisingly, our MCTF even performs better than the `full' base model (\textcolor{red}{red} dotted line) with a reduced computational complexity. 
Specifically, it brings a 0.5\%, and 0.3\% gain while reducing FLOPs by about 44\% in DeiT-T, and DeiT-S~\cite{touvron2021training}, respectively. We have observed a similar speed-up (31\%) in T2T-ViT~\cite{yuan2021tokens}, and LV-ViT~\cite{jiang2021all} without any performance degradation.\\
\noindent Our contributions are summarized in fourfold. 
\begin{itemize}
    \item We propose \textit{Multi-criteria Token Fusion}, a novel token fusion method that considers multi-criteria, \eg, similarity, informativeness, and size, to capture the complex relationship of tokens and minimize information loss.
    \item For measuring the informativeness of the tokens, we utilize \emph{one-step-ahead attention} to retain the attentive tokens in the following layers.
    \item We propose a new fine-tuning scheme with \emph{token reduction consistency} to boost the generalization performance of transformers equipped with MCTF.
    \item The extensive experiments demonstrate that MCTF achieves the best speed-accuracy trade-off in diverse ViTs, surpassing all previous token reduction methods.     
\end{itemize}

%% file: Figures/Figure1.tex
\begin{figure*}[t]
     \centering
     \begin{subfigure}[b]{0.48\linewidth}
         \centering
         \includegraphics[trim=0 0 0 0,clip, width=\linewidth]{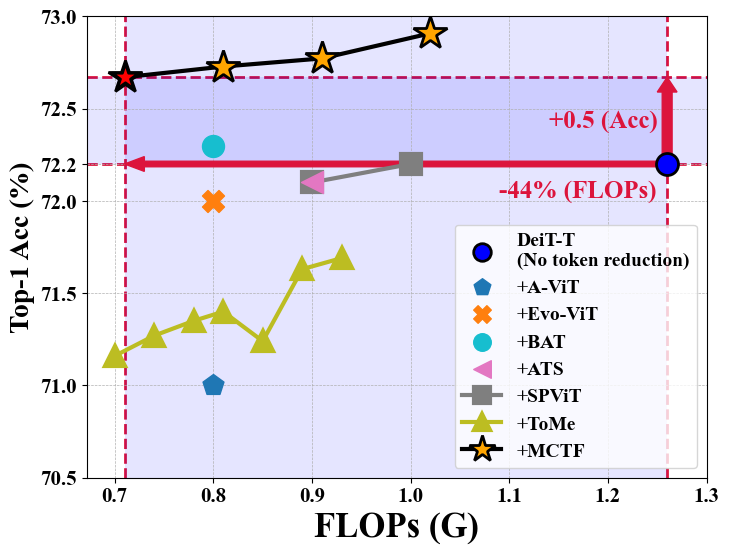}
     \end{subfigure}
     \hfill
     \begin{subfigure}[b]{0.48\linewidth}
         \centering
         \includegraphics[trim=0 0 0 0,clip, width=\linewidth]{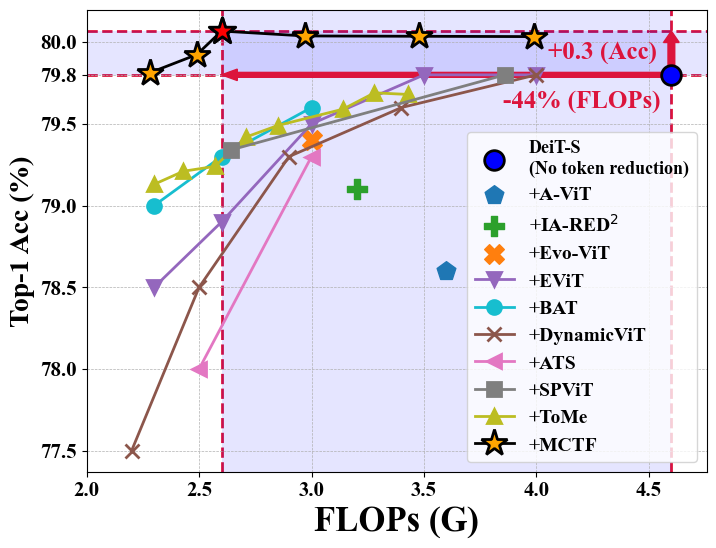}
     \end{subfigure}
     \caption{\textbf{Comparison of the token reduction methods with DeiT-T (left), and DeiT-S (right).}
    Given a base model marked as \blue{blue} circle, previous token reduction methods accelerate the speed with the trade-off between accuracy and computational cost. Our MCTF, marked as a star, even brings performance improvements while lessening the complexity of DeiT. Note that after only one finetuning with the specific reduced number of tokens marked as \red{red} star, we simply evaluate it with the diverse FLOPs by adjusting the reduced numbers.}
     \label{fig:comparison}
\vspace{-10pt}     
\end{figure*}

%% file: section/relatedworks.tex
\section{Related works}
\label{sec:5}
\paragraph{Vision Transformers.} Vision Transformer \cite{dosovitskiy2020image} is introduced to tackle the vision tasks. Later, DeiT~\cite{touvron2021training} and CaiT~\cite{touvron2021going} are proposed to handle the data efficiency and scalability of ViT, respectively. Recent works~\cite{liu2021swin, wang2021pyramid, chen2021crossvit, heo2021rethinking,dong2022cswin} tried to insert the inductive biases of CNNs on ViT, such as the locality or pyramid-architecture. In parallel, there is a line of works that boosts the vanilla ViT by scaling~\cite{touvron2021going, zhai2022scaling} or self-supervised learning~\cite{he2022masked, bachmann2022multimae, wudenoising}. Despite the promising results of these works, the quadratic complexity of ViTs is still the major constraint for scaling the model. For the sake of mitigating the complexity, Reformer~\cite{kitaev2020reformer} lessens the quadratic complexity to $O(N\log N)$ through the hashing function, and Linformer~\cite{wang2020linformer}, performer~\cite{choromanski2020rethinking}, and Nystr\"omformer~\cite{xiong2021nystromformer} achieve the linear cost with the approximated linear attention. Also, several works~\cite{liu2021swin,chu2021twins, arar2022learned, dong2022cswin} utilize sparse attention with the reduced key or query. Swin~\cite{liu2021swin} and Twins~\cite{chu2021twins} utilize the local attention within the fixed size of the window to mitigate the complexity.


\paragraph{Token reduction in ViTs.}
Most of the computational burden in ViTs arises from the self-attention. To reduce the quadratic cost in the number of tokens, recent works~\cite{meng2022adavit,fayyaz2022adaptive,yin2022vit,rao2021dynamicvit,pan2021ia,kong2022spvit,liang2022not,marin2023token,bolya2022token,long2023beyond} have an interest in reducing the token itself. These works have the advantage of utilizing the original ViTs architecture without modification. 
In earlier works~\cite{meng2022adavit,fayyaz2022adaptive,yin2022vit,rao2021dynamicvit,pan2021ia}, the uninformative tokens are simply dropped during the forward process, leading to the information loss. To compensate for this, SPViT~\cite{kong2022spvit} and EViT~\cite{liang2022not} first split the tokens into informative and uninformative token sets based on attention scores, then fuse these uninformative token sets into a single token. In parallel, token pooling~\cite{marin2023token} and ToMe~\cite{bolya2022token} combine the semantically similar tokens to reduce redundancies. A more recent study BAT~\cite{long2023beyond} first split the tokens based on informativeness then fuse the tokens considering the diversity of the tokens. Despite the advantage of each criterion, successful integration of multi-criteria is still less explored.

%% file: section/method.tex
\input{Figures/Figure2}
\section{Method}
\label{sec:3}

We first review the self-attention and token reduction approaches (\Cref{sec:3.1}).
Then, we present our multi-criteria token fusion (\Cref{sec:3.2}) that leverages one-step-ahead attention (\Cref{sec:3.3}). Lastly, we introduce a training strategy with token reduction consistency in~\Cref{sec:3.4}.

\subsection{Preliminaries}
\label{sec:3.1}
In Transformers, tokens $\mathbf{X} \in \mathbb{R}^{N\times C}$ are processed by self-attention defined as 
\begin{align}
\text{SA}(\mathbf{X}) = \text{softmax}\left(\frac{\mathbf{QK}^\top}{\sqrt{C}} \right) \mathbf{V}, 
\label{eq:SA}
\end{align}
where $\mathbf{Q, K, V} = \mathbf{XW_Q, XW_K,XW_V}$, and $\mathbf{W_Q, W_K,}$ $\mathbf{W_V} \in \mathbb{R}^{C\times C}$ are learnable weight matrices. 
Despite its outstanding expressive power, the self-attention does not scale well with the number of tokens $N$ due to its quadratic time complexity $O(N^2 C + N C^2)$.
To address this problem, 
a line of works~\cite{meng2022adavit,fayyaz2022adaptive,yin2022vit,rao2021dynamicvit, pan2021ia} reduces the number of tokens simply by \textit{pruning} uninformative tokens.
These approaches often cause significant performance degradation due to the loss of information.
Thus, another line of works~\cite{kong2022spvit, liang2022not,marin2023token,bolya2022token,long2023beyond} \textit{fuses} the uninformative or redundant tokens $\hat{\mathbf{X}} \subset \mathbf{X}$ into a new token $\hat{\mathbf{x}} = \delta(\hat{\mathbf{X}})$, where $\mathbf{X}$ is the set of original tokens, and $\delta$ denotes a merging function, \eg, max-pooling or averaging.
In this work, we also adopt `token fusion' rather than `token pruning' with multiple criteria to minimize the loss of information by token reduction.
\input{Figures/Figure3}
\subsection{Multi-criteria token fusion}
\label{sec:3.2}
Given a set of input tokens $\mathbf{X} \in \mathbb{R}^{N\times C}$, the goal of MCTF is to fuse the tokens into output tokens $\hat{\mathbf{X}} \in \mathbb{R}^{(N-r)\times C}$, where $r$ is the number of fused tokens. 
To minimize the information loss, we first evaluate the relations between the tokens based on multi-criteria, then group and merge the tokens through bidirectional bipartite soft matching.\\
\textbf{Multi-criteria attraction function.}
We first define an attraction function $\mathbf{W}$ based on multiple criteria as
\begin{align}
\mathbf{W}(\mathbf{x}_i, \mathbf{x}_j) = \Pi^M_{k=1}(\mathbf{W}^k(\mathbf{x}_i, \mathbf{x}_j))^{\tau_k},
\end{align}
where $\mathbf{W}^k:\mathbb{R}^C \times \mathbb{R}^C \rightarrow \mathbb{R}_{+}$ is an attraction function computed by $k$-th criterion, and $\tau^k \in \mathbb{R}_{+}$ is the temperature parameter to adjust the influence of $k$-th criterion.
The higher attraction score between two tokens indicates a higher chance of being fused.
In this work, we consider the following three criteria: similarity, informativeness, and size.\\
\textbf{Similarity.} The first criterion is the similarity of tokens to reduce redundant information. 
Akin to the previous works~\cite{marin2023token, bolya2022token} requiring the proximity of tokens, we leverage the cosine similarity between the set of tokens for
\begin{align}
\mathbf{W}^\text{sim}(\mathbf{x}_i, \mathbf{x}_j) = \frac{1}{2} \left ( \frac{\mathbf{x}_i \cdot \mathbf{x}_j}{\|\mathbf{x}_i\|\|\mathbf{x}_j\|}+1 \right ).
\label{eq:w_sim}
\end{align}
Token fusion with similarity effectively eliminates the redundant tokens, yet it often excessively combines the informative tokens as in~\Cref{fig:2.b}, causing the loss of information.\\
\textbf{Informativeness.} To minimize the information loss, we introduce informativeness to avoid the fusion of informative tokens.
To quantify the informativeness, we measure the averaged attention scores $\mathbf{a} \in [0,1]^N$ in the self-attention layer, which indicates the impact of each token on others: $\mathbf{a}_j = \frac{1}{N}\sum^N_i \mathbf{A}_{ij}$, where $\mathbf{A}_{ij} = \text{softmax}\left(\frac{\mathbf{Q}_i\mathbf{K}_j^T}{\sqrt{C}}\right)$. 
When $\mathbf{a}_i \rightarrow 0$, there's no influence from $\mathbf{x}_i$ to other tokens.
With the informativeness scores, we define an informativeness-based attraction function as
\begin{align}
    \mathbf{W}^\text{info}(\mathbf{x}_i, \mathbf{x}_j) = \frac{1}{\mathbf{a}_i \mathbf{a}_j},
    \label{eq:w_info}
\end{align}
where $\mathbf{a}_i, \mathbf{a}_j$ are the informative scores of $\mathbf{x}_i, \mathbf{x}_j$, respectively. 
When both tokens are uninformative ($\mathbf{a}_i, \mathbf{a}_j \rightarrow 0$), 
the weight gets higher ($\mathbf{W}^\text{info}(\mathbf{x}_i, \mathbf{x}_j) \rightarrow \infty$), making two tokens prone to be fused. 
In~\Cref{fig:2.c}, with the weights combined with the similarity and informativeness, the tokens in the foreground object are less fused. \\
\textbf{Size.} The last criterion is the size of the tokens, which indicates the number of fused tokens. 
Although tokens are not dropped but merged via a merging function, \eg, averaging pooling or max pooling, 
it is difficult to preserve all the information as the number of constituent tokens increases. 
So, the fusion between smaller tokens is preferred.
To this end, we initially set the size $\mathbf{s} \in \mathbb{N}^N$ of tokens $\mathbf{X}$ as 1 and track the number of constituent (fused) tokens of each token, and define a size-based attraction function as
\begin{align}
    \mathbf{W}^\text{size}(\mathbf{x}_i, \mathbf{x}_j) = \frac{1}{\mathbf{s}_i \mathbf{s}_j}.
    \label{eq:w_size}
\end{align}
In~\Cref{fig:2.d}, tokens are merged based on the multi-criteria: similarity, informativeness, and size. 
We observed that the fusion happens between similar tokens and 
the fusion of foreground tokens or large tokens is properly suppressed. 

\paragraph{Bidirectional bipartite soft matching.}
Given the multi-criteria-based attraction function $\mathbf{W}$, our MCTF performs a \emph{relaxed} bidirectional bipartite matching called bipartite soft matching~\cite{bolya2022token}. 
One advantage of bipartite matching is that it alleviates the quadratic cost of similarity computation between tokens, \ie, $O(N^2) \rightarrow O(N'^{2})$, where $N^\prime = \lfloor \frac{N}{2} \rfloor$. 
In addition, by relaxing the one-to-one correspondence constraints, the solution can be obtained by an efficient algorithm.
In this relaxed matching problem, the set of tokens $\mathbf{X}$ is first split into the source and target $\mathbf{X}^\alpha, \mathbf{X}^\beta \in \mathbb{R}^{N^\prime \times C}$ as in Step 1 of~\Cref{fig:bi}.
Given a set of binary decision variables, \ie, the edge matrix $\mathbf{E} \in \{0,1\}^{N' \times N'}$ between $\mathbf{X}^\alpha, \text{ and } \mathbf{X}^\beta$,
bipartite soft matching is formulated as 
\begin{align}
    \label{eq:obj}
    \mathbf{E}^\ast=&\argmax_{\mathbf{E}} {\sum}_{ij} \mathbf{w}^\prime_{ij} \mathbf{e}_{ij}\\  
    &\text{subject to} {\sum}_{ij} \mathbf{e}_{ij} = r, \;{\sum}_{j} \mathbf{e}_{ij} \leq 1\;\forall{i},
\end{align}
where
\begin{align}
    \mathbf{w}^\prime_{ij} = \begin{cases}
    \mathbf{w}_{ij} &\text{if } j \neq \argmax_{j^\prime} \mathbf{w}_{ij^\prime}\\
    0 & \text{otherwise}
    \end{cases}, 
\end{align}
$\mathbf{e}_{ij}$ indicates the presence of the edge between $i,j$-th token of $\mathbf{X}^\alpha, \mathbf{X}^\beta$, and , 
    $\mathbf{w}_{ij} = \mathbf{W}(\mathbf{x}^\alpha_i, \mathbf{x}^\beta_j)$.
This optimization problem can be solved by two simple steps: 1) find the best edge that maximizes $\mathbf{w}_{ij}$ for each $i$, and 
2) choose the top-$r$ edges with the largest attraction scores. 
Then, based on the soft matching result $\mathbf{E}^\ast$, we group the tokens as
\begin{align}
&\mathbf{X}^{\alpha \rightarrow \beta}_j = \{\mathbf{x}^\alpha_i \in \mathbf{X}^\alpha \; | \; \mathbf{e}_{ij} =1\} \cup \{\mathbf{x}^\beta_j\},
\end{align}
where $\mathbf{X}^{\alpha\rightarrow\beta}_i$ indicates the set of tokens matched with $\mathbf{x}^\beta_i$.
Finally, the results of the fusion $\tilde{\mathbf{X}}$ are obtained as
\begin{align} 
\tilde{\mathbf{X}} = \tilde{\mathbf{X}}^\alpha &\cup \tilde{\mathbf{X}}^\beta, \label{eq:MCTF_1}\\
\text{where }\tilde{\mathbf{X}}^\alpha &=  \mathbf{X}^\alpha - {\bigcup}^{N^\prime}_i \mathbf{X}^{\alpha \rightarrow \beta}_i, \label{eq:MCTF_2}\\
\tilde{\mathbf{X}}^\beta &= {\bigcup}^{N^\prime}_i \{\delta(\mathbf{X}^{\alpha \rightarrow \beta}_i)\}, \label{eq:MCTF_3}
\end{align}
$\delta(\mathbf{X}) = \delta(\{\mathbf{x}_i\}_i)= \sum_i \frac{\mathbf{a}_i\mathbf{s}_i\mathbf{x}_i}{\sum_{i^\prime} \mathbf{a}_{i^\prime}\mathbf{s}_{i^\prime}}$ is the pooling operation considering the attention scores $\mathbf{a}$ and the size $\mathbf{s}$ of the tokens.
Still, as shown in Step2 of~\Cref{fig:bi}, the number of target tokens $\mathbf{X}^\beta$ cannot be reduced.
To handle this issue, MCTF performs \textit{bidirectional bipartite soft matching} by conducting the matching in the opposite direction with the updated token sets $\tilde{\mathbf{X}}^\alpha$, and $\tilde{\mathbf{X}}^\beta$ as in Step 3, 4 of ~\Cref{fig:bi}.
The final output tokens $\hat{\mathbf{X}} = \hat{\mathbf{X}}^\alpha \cup \hat{\mathbf{X}}^\beta $ are defined with the following.
\begin{align}
    &\hat{\mathbf{X}}^\alpha =  {\bigcup}^{N^\prime-r}_i \{\delta(\tilde{\mathbf{X}}^{\beta \rightarrow \alpha}_i)\}, \label{eq:MCTF_4}\\
    &\hat{\mathbf{X}}^\beta  = \tilde{\mathbf{X}}^\beta - {\bigcup}^{N^\prime - r}_i \tilde{\mathbf{X}}^{\beta \rightarrow \alpha}_i. \label{eq:MCTF_5}
\end{align}
Note that calculating the pairwise weights with updated two sets of tokens $\tilde{\mathbf{w}}_{ij} = \mathbf{W} (\tilde{\mathbf{x}}^\beta_i, \tilde{\mathbf{x}}^\alpha_j)$ introduces the additional computational costs of $O(N^\prime (N^\prime - r))$. To avoid this overhead, we approximate the attraction function by the attraction scores before fusion.
In short, we just reuse the pre-calculated weights since $\tilde{\mathbf{X}}^\alpha$ is the subset of $\mathbf{X}^\alpha$.
This allows MCTF to efficiently reduce tokens considering bidirectional relations between two subsets with negligible extra costs compared to uni-directional bipartite soft matching.

\input{Figures/Figure4}
\subsection{One-step-ahead attention for informativeness}
\label{sec:3.3}
In assessing informativeness, prior works~\cite{kong2022spvit,liang2022not,long2023beyond} have leveraged the attention scores from the previous self-attention layer.
As illustrated in Figure~\ref{fig:oaa}, previous approaches use the attention $\textbf{A}^l$ from the previous layer to fuse tokens $\mathbf{X}^{l}$. 
This technique allows efficient assessment under the assumption that the attention maps in consecutive layers are similar. 
However, we observed that the attention maps often substantially differ, as shown in~\Cref{fig:att_diff}, and the attention from a previous layer may lead to suboptimal token fusion. 
Thus, we proposed \textbf{one-step-ahead attention}, which measures the informativeness of tokens based on the attention map in the next layer, \ie, $\mathbf{A}^{l+1}$.
Then, the informativeness scores $\mathbf{a}$ in~\Cref{eq:w_info} is calculated with $\mathbf{A}^{l+1}\in \mathbb{R}^{N \times N}$.
This simple remedy provides a considerable improvement; see~\Cref{fig:abl2} in~\Cref{sec:4.2}.
After token fusion, we efficiently compute the attention map $\hat{\mathbf{A}}^{l+1} \in  \mathbb{R}^{(N-r)\times (N-r)}$ of fused tokens $\hat{\mathbf{X}}^l\in \mathbb{R}^{(N-r) \times C}$
by simply aggregating $\mathbf{A}^{l+1} \in \mathbb{R}^{N\times N}$
without recomputing the dot-product self-attention. 
To be specific, when the tokens are fused as $\delta(\{\mathbf{x}_i\}_i)$ during~\Cref{eq:MCTF_1,eq:MCTF_2,eq:MCTF_3,eq:MCTF_4,eq:MCTF_5}, their corresponding one-step-ahead attention scores are also fused as $\delta(\{\mathbf{A}^{l+1}_i\}_i)$ in both query and key direction. 
Note that when fusing attention scores for queries we use simple sum for $\delta$,\ie, $\forall_i \sum_j \hat{\mathbf{A}}^{l+1}_{ij}=1$.   
For fusing attention scores for queries, we use simple sum for $\delta$ to guarantee $\forall_i \sum_j \hat{\mathbf{A}}^{l+1}_{ij}=1$.

\input{Figures/Figure5}
\subsection{Token reduction consistency}
\label{sec:3.4}
We here propose a new fine-tuning scheme to further improve the performance of vision Transformer $f_\theta(\cdot;r)$ with MCTF.
We observe that a different number of reduced tokens per layer, denoted as $r$, may lead to different representations of samples. 
By training Transformers with different $r$ and encouraging the consistency between them, namely, token reduction consistency, 
we achieve the additional performance gain. 
The objective function of our method is given as
\begin{align}
    \mathcal{L} = \mathcal{L}_\text{CE}(f_\theta(x;r), y) &+ \mathcal{L}_\text{CE}(f_\theta(x;r^\prime), y) \nonumber \\
    &+  \lambda\mathcal{L}_\text{MSE}(\mathbf{x}^\text{cls}_r, \mathbf{x}^\text{cls}_{r^\prime}),
    \label{eq:con}
\end{align}
where $(x,y)$ is a supervised sample, $r, r^\prime$ is the fixed and dynamic reduced token numbers, $\lambda$ is the coefficient for consistency loss, and $\mathbf{x}^\text{cls}_r, \mathbf{x}^\text{cls}_{r^\prime}$ are the class tokens in the last layer of models $f_\theta(x;r), f_\theta(x;r^\prime)$.
In this objective, we first calculate the cross-entropy loss $\mathcal{L}_\text{CE}(f_\theta(x;r), y)$ with fixed $r$, which is the target reduction number that will be used in the evaluation. 
At the same time, we generate another representation of the input $x$ with smaller but randomly drawn $r^\prime \sim \text{uniform}(0,r)$, and calculate the loss $\mathcal{L}_\text{CE}(f_\theta(x;r^\prime), y)$. 
Then, we impose the token consistency loss $\mathcal{L}_\text{MSE}(\mathbf{x}^\text{cls}_r, \mathbf{x}^\text{cls}_{r^\prime})$ on the class tokens, 
to retain the consistent representation across the diverse reduced token numbers $r^\prime$.
The proposed method can be viewed as a new type of token-level data augmentation~\cite{liu2023tokenmix,choi2022tokenmixup} and consistency regularization. 
Our token reduction consistency encourages the representation $\xclsr$ obtained by the target reduction number $r$ to mimic the slightly augmented representation 
$\mathbf{x}^\text{cls}_{r^\prime}$, which is more similar to ones with no token reduction since $r^\prime < r$.

%% file: Figures/Figure2.tex
\begin{figure*}[t]
     \centering
     \begin{subfigure}[b]{0.22\linewidth}
         \centering
         \includegraphics[trim=0 0 0 0,clip, height=\linewidth]{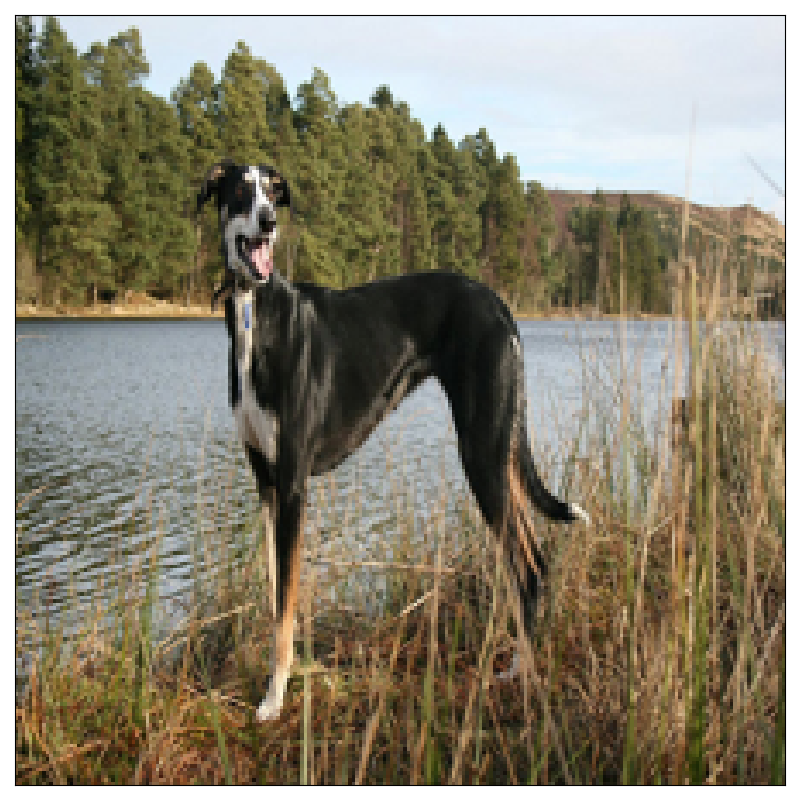}
         \caption{Origin}
         \label{fig:2.a}
     \end{subfigure}
     \hfill
     \begin{subfigure}[b]{0.22\linewidth}
         \centering
         \includegraphics[trim=0 0 0 0,clip, height=\linewidth]{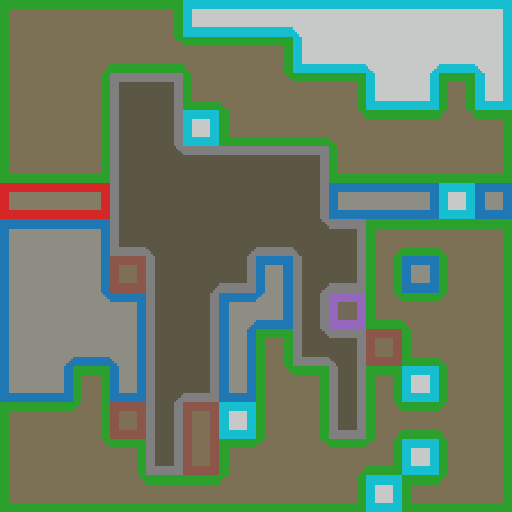}
         \caption{$\mathbf{W^\text{sim}}$}
         \label{fig:2.b}
     \end{subfigure}
     \hfill
     \begin{subfigure}[b]{0.22\linewidth}
         \centering
         \includegraphics[trim=0 0 0 0,clip, height=\linewidth]{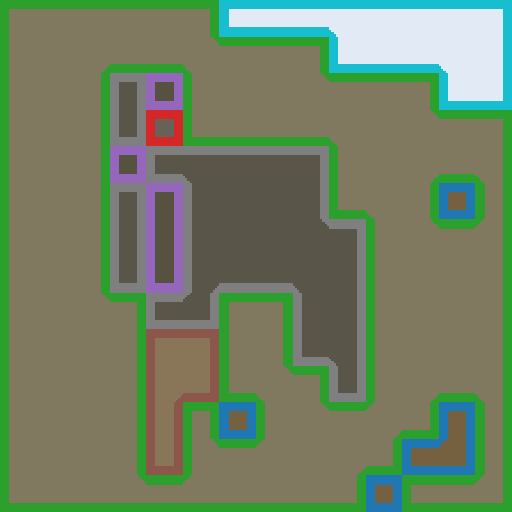}
         \caption{$\mathbf{W^\text{sim}} \& \mathbf{W^\text{info}}$}
         \label{fig:2.c}
     \end{subfigure}
     \hfill
     \begin{subfigure}[b]{0.22\linewidth}
         \centering
         \includegraphics[trim=0 0 0 0, clip, height=\linewidth]{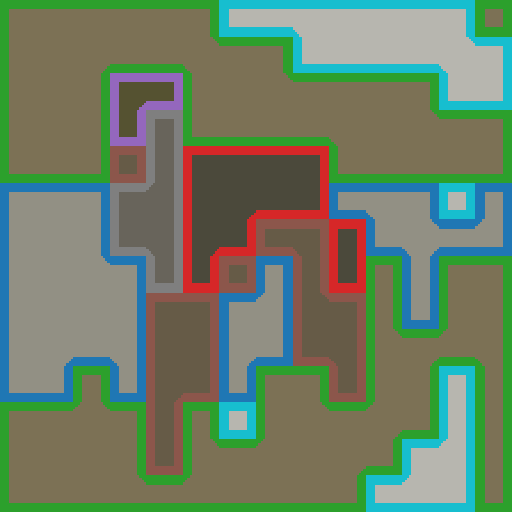}
         \caption{$\mathbf{W^\text{sim}} \& \mathbf{W^\text{info}} \& \mathbf{W^\text{size}}$}
         \label{fig:2.d}
     \end{subfigure}
     
    \caption{\textbf{Visualization of the fused tokens.} Given (a) the leftmost image, (b) fusing the tokens with a single criterion $\textbf{W}^\text{sim}$ often results in the excessive fusion of the foreground object. (c) Then considering both similarity and informativeness ($\textbf{W}^\text{sim}\&\textbf{W}^\text{info}$), tokens in the foreground objects are less fused while the tokens in the background are largely fused. (d) Finally, MCTF helps retain the information of each component in the image by preventing the large-size token with the multi-criteria ($\textbf{W}^\text{sim}\&\textbf{W}^\text{info}\&\textbf{W}^\text{size})$.}
    \label{fig:mc}
\end{figure*}

%% file: Figures/Figure3.tex
\begin{figure*}[t]
     \centering
     \includegraphics[trim=0 0 0 0,clip,width=0.9\textwidth]{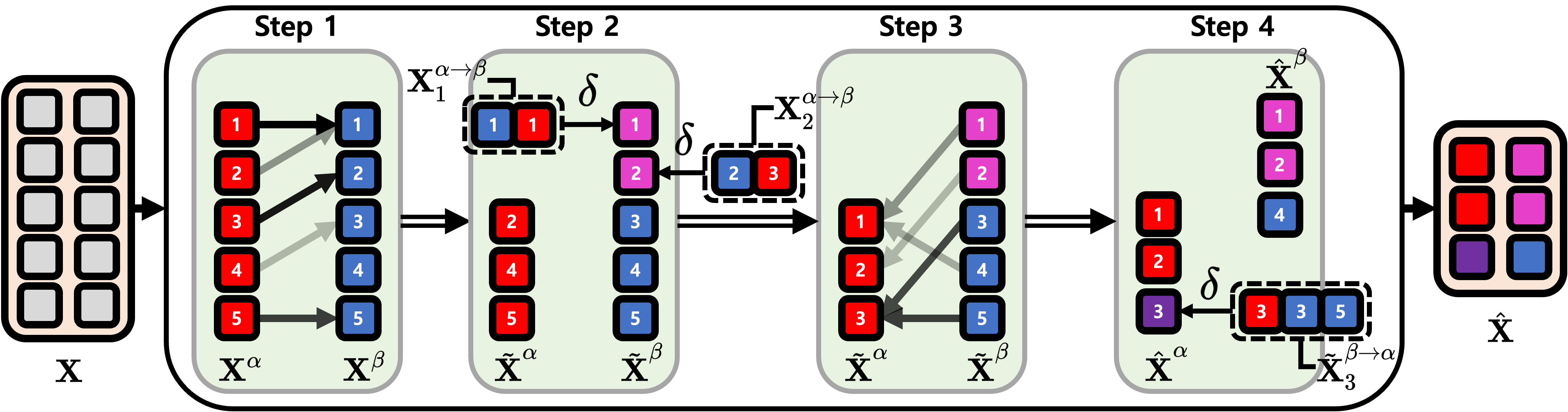}
    \caption{\textbf{Bidirectional bipartite soft matching.} The set of tokens $\mathbf{X}$ is split into two groups $\mathbf{X}^\alpha, \mathbf{X}^\beta$, and bidirectional bipartite soft matching are conducted through Step 1-4. The intensity of the lines indicates the multi-criteria weights $\mathbf{W}^t$.}
    \label{fig:bi}
\vspace{-10pt}
\end{figure*}

%% file: Figures/Figure4.tex
\begin{figure}[t]
    \centering
    \includegraphics[trim=0 0 0 0,clip,width=\linewidth]{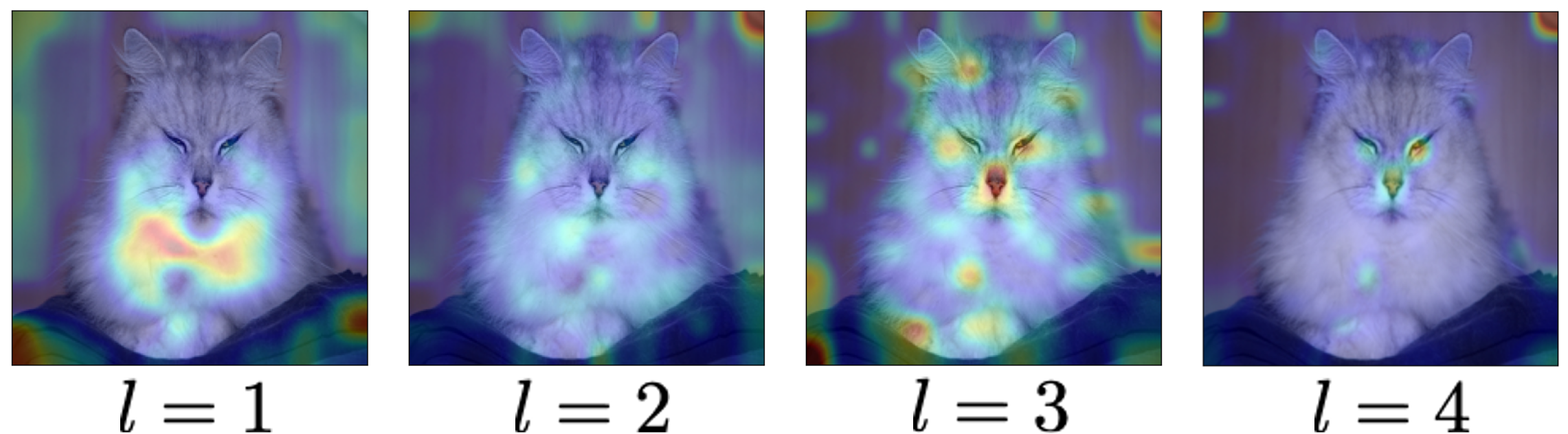}
    \caption{\textbf{Visualization of attentiveness in consecutive layers.}}
    \label{fig:att_diff}
\vspace{-10pt}
\end{figure}
\begin{figure}[t]
    \centering
    \includegraphics[trim=0 0 0 0,clip,width=\linewidth]{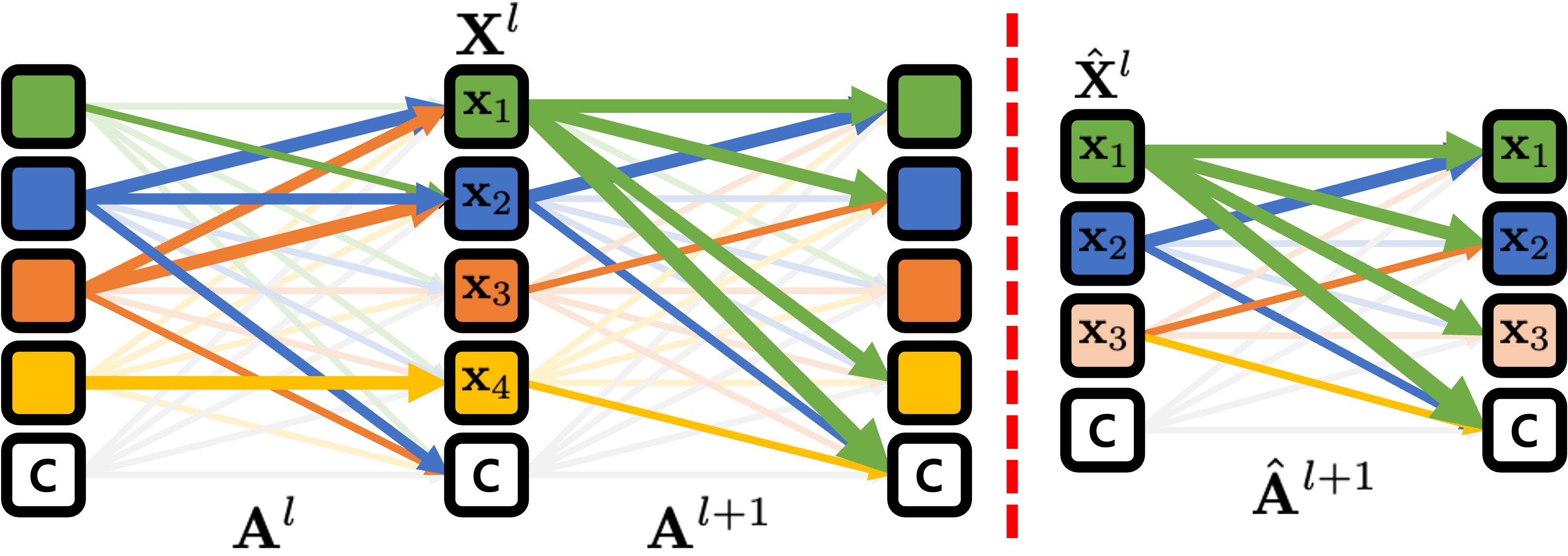}
    \caption{\textbf{Illustration of attention map in the consecutive layers and approximated attention.} (Left) The attention score $\mathbf{A}^l$ is the past influence of the tokens to generate $\mathbf{X}^l$.
    If we fuse the tokens $\mathbf{X}^l$ based on $\mathbf{A}^l$, $\mathbf{x}_1$ is prone to be fused despite the highest informativeness score in the following attention. So, we instead leverage the informativeness based on the one-step-ahead attention $\mathbf{A}^{l+1}$. (Right) After the fusion, we also aggregate the $\mathbf{A}^{l+1}$ to approximate the attention map $\hat{\mathbf{A}}^{l+1}$ for updating fused tokens $\hat{\mathbf{X}}^l$.}
    \label{fig:oaa}
\vspace{-10pt}
\end{figure}


%% file: Figures/Figure5.tex
\begin{figure}[t]

    
    

    


    
    \centering
    \includegraphics[trim=0 0 0 0,clip,width=0.47\textwidth]{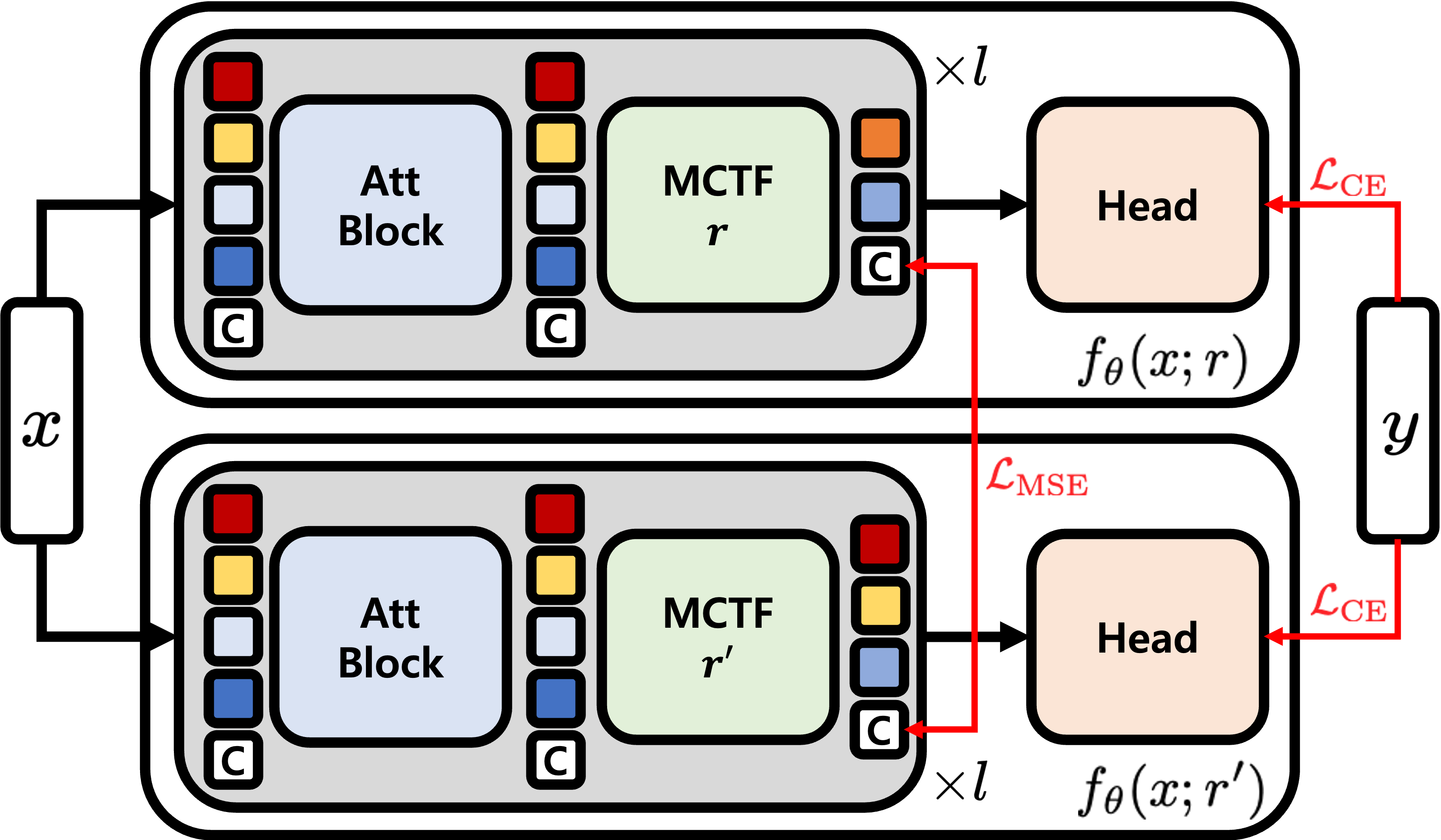}
        \caption{\textbf{Illustration of training with token reduction consistency.} During training, we forward the input $x$ as $f(x;r)$, and $f(x; r^\prime)$, respectively. To obtain the augmented representation, $r^\prime$ is randomly selected in every step, and the model is updated with supervisory signals $\mathcal{L}_\text{CE}$, and consistency loss $\mathcal{L}_\text{MSE}$.}
    \label{fig:SD}
\vspace{-10pt}
\end{figure}

%% file: section/experiments.tex
\input{tables/table_main1}
\section{Experiments}
\label{sec:4}
\textbf{Baselines.}
To validate the effectiveness of the proposed methods, we compare MCTF with the previous token reduction methods. For comparison, we opt the token pruning methods (A-ViT~\cite{yin2022vit}, IA-RED$^2$~\cite{pan2021ia}, DynamicViT~\cite{rao2021dynamicvit}, EvoViT~\cite{xu2022evo}, ATS~\cite{fayyaz2022adaptive}) and token fusion methods (SPViT~\cite{kong2022spvit}, EViT~\cite{liang2022not}, ToMe~\cite{bolya2022token}, BAT~\cite{long2023beyond}) in DeiT~\cite{touvron2021training}, and report the efficiency (FLOPs (G)) and the performance (Top-1 Acc (\%)) of each method. 
Further, to validate MCTF on other Vision Transformers (T2T-ViT~\cite{yuan2021tokens}, LV-ViT~\cite{jiang2021all}), we report the results of MCTF and compare it with the official number of existing works.
We denote the number of reduced tokens per layer $r$ with the subscript in~\Cref{tab:main1,tab:main2}.
The gray color in the table indicates the base model, and the green and red color indicates the improvements and degradations of the performance compared to the base model, respectively.

\input{tables/table_main2}
\input{Figures/Figure6}
\subsection{Experimental Results}
\label{sec:4.1}
\textbf{Comparison of the token reduction methods.} The comparison with existing token reduction methods is summarized in~\Cref{tab:main1}. 
We demonstrate that our MCTF achieves the best performance with the lowest FLOPs in DeiT~\cite{touvron2021training} surpassing all previous works. 
Further, it is worth noting that MCTF is the only work that avoids performance degradation with the lowest FLOPs in both DeiT-T and DeiT-S.
Through Finetuning DeiT-T for 30 epochs, MCTF brings a significant gain of +0.5\% in accuracy over the base model with nearly half FLOPs. Similarly, we observe a gain of +0.3\% with DeiT-S while boosting the FLOPs by -2.0 (G). 
We believe that multi-criteria with one-step-ahead attention helps the model to minimize the loss of information; further consistency loss on the class token through the token reduction improves the generalizability of the model.

\noindent\textbf{MCTF with other Vision Transformers.} 
To validate the applicability of MCTF in various ViTs, we demonstrate MCTF with other transformer architectures in~\Cref{tab:main2}. Following previous works~\cite{liang2022not,long2023beyond,rao2021dynamicvit,kong2022spvit}, we apply MCTF with LV-ViT. Also, we present the results of MCTF with T2T-ViT. As presented in the table, our experimental results are promising. MCTF in these architectures gets at least 31\% speedup without performance degradation. Further, MCTF combined with LV-ViT outperforms all other Transformers and token reduction methods regarding FLOPs, and accuracy.
Especially, it is worth noting that all token reduction methods except for MCTF bring the performance degradation in LV-ViT.
These results reveal that MCTF is the efficient token reduction method for the diverse Vision Transformers. \input{tables/table_main3}

\noindent\textbf{Token reduction without training.} 
Similar to ToMe~\cite{bolya2022token}, MCTF is applicable with pre-trained ViTs without any additional training since MCTF does not require any learnable parameters.
We here apply the two reduction methods to the pre-trained DeiT without finetuning and provide the results in~\Cref{tab:main3}. 
Regardless of the reduced number of tokens  $r$ in each layer, MCTF consistently surpasses ToMe. Especially, in the most sparse setting $r=20$, the performance gap is significant (+7.0\% in DeiT-T, +3.8\% in DeiT-S).
Note that without any additional training, our MCTF$_{r=16}$ with pre-trained DeiT-S still shows a competitive performance of 79.2\% compared to reduction methods requiring training (\eg., 78.6\% of A-ViT, 79.1\% of IA-RED$^2$, and 79.3\% of DynamicViT, and SPViT in~\Cref{tab:main1}).


\subsection{Ablation studies on MCTF}
\label{sec:4.2}
We provide ablation studies to validate each component of MCTF. 
Unless otherwise stated, we conduct whole experiments with DeiT-S finetuned with MCTF ($r=16$). We provide the FLOPs-Accuracy graph by adjusting the reduced number of tokens per layer $r \in [1,20]$.

\input{Figures/Figure7}
\noindent\textbf{Multi-criteria.} We explore the effectiveness of multi-criteria in~\Cref{fig:abl1}. 
First, regarding the multi-criteria, we utilize three criteria for MCTF, \textit{i.e.}, similarity (\textbf{sim.}), informativeness (\textbf{info.}), and \textbf{size}. Each single criterion of similarity and informativeness shows a relatively inferior performance compared to dual (sim. \& info.) and multi-criteria (sim. \& info. \& size). Specifically, when $r=16$, the performance of a single criterion is 79.7\%, and 79.4\% with similarity and informativeness, respectively. Then, adopting dual criteria (sim. \& info.), MCTF achieves 79.8\%. Finally, we get an accuracy of 80.1\% with a gain of +0.3\% by respecting all three criteria (sim. \& info. \& size).
These performance gaps get larger as $r$ increases, which proves the importance of the multi-criteria for token fusion. 

\noindent\textbf{One-step-ahead attention and token reduction consistency.}
To show the validity of one-step-ahead attention and token reduction consistency, we also provide the results of MCTF with and without each component in~\Cref{fig:abl2}. When eliminating either one-step-ahead attention or token reduction consistency, the accuracies are dropped in every FLOP. This significant drop indicates that both approaches matter for MCTF.
In short, by adopting one-step-ahead attention and token reduction consistency, MCTF effectively mitigates the performance degradation in a wide range of FLOPs.

\noindent\textbf{Comparison of design choices.} The ablations on design choices are presented in~\Cref{tab:abl1}. First, our \textbf{bidirectional} bipartite matching, which enables capturing the bidirectional relation in two sets, enhances the accuracy compared to \textbf{one-way} bipartite matching. Next, for pooling operation $\delta$, the weighted sum considering the size $\mathbf{s}$ and attentiveness $\mathbf{a}$ is a better choice than others like max-pool or average. Lastly, we compare the results with the precise and approximated attention for $\hat{\mathbf{A}}^l$. For precise attention, we just conduct the similarity calculation for one-step-ahead attention and the attention in the self-attention layer after fusion, separately. Otherwise, we approximate it with one-step-ahead attention as described in~\Cref{sec:3.3}. As presented in the table, our approximated attention maintains the performance with the substantial improvement in efficiency (-0.4 (G) FLOPs).
\input{tables/table_abl1}

\subsection{Analyse of MCTF}
\noindent\textbf{Qualitative results.} For a better understanding of MCTF, we provide the qualitative results of MCTF in~\Cref{fig:vis}. 
We visualize the fused tokens at the last block of DeiT-S on ImageNet-1K and denote the fused tokens by the same border color. As shown in the figure, since the tokens are merged with multi-criteria (\eg, similarity, informativeness, size), we maintain the more diverse tokens in the informative foreground object. For instance, in the third image of the hamster, while the background patches including the hand are fused into one token, the foreground tokens are less fused while maintaining the details like the eye, ear, and face of the hamster.
In short, compared to the background, the foreground tokens are less fused with the moderate size retaining the information of the main content.

\noindent\textbf{Soundness of size criterion.} \Cref{fig:hist} presents the histogram of sizes of tokens after token reduction with and without size criterion. 
Specifically, we measure the size of the largest token at the last block and provide the histogram.  
With our size criterion, the merged tokens tend to have smaller sizes s showing the average size of 39.3/49.2 with and without the Size criterion, respectively. 
As intended, MCTF successfully suppresses the large-sized tokens, which are a source of information loss, leading to performance improvement.

\input{Figures/Figure8}

%% file: tables/table_main1.tex
\begin{table}[t!]
  \centering 
  \caption{\textbf{Image classification results}}
  \label{tab:main1} 
  \vspace{-5pt}
  \setlength{\tabcolsep}{1.9pt}
  \begin{tabular}{lcccc}
  \toprule
  \multicolumn{1}{c}{\multirow{2}{*}{Method}}&   FLOPs & Params  & Top-1 Acc  \\
  &(G)&(M)& (\%)\\ 
  \midrule
  
  \rowcolor{Gray}DeiT-T~\cite{touvron2021training}  &         1.2  & 5  &   72.2 (-)             \\
  \hspace{2pt}+EvoViT\gray{$_\text{[AAAI '22]}$}~\cite{xu2022evo}   &        0.8  & 5  &72.0 (\red{-0.2})    \\
  \hspace{2pt}+A-ViT\gray{$_\text{[CVPR '22]}$}~\cite{yin2022vit}                   &         0.8  &  5  &       71.0 (\red{-1.2})    \\ 
  \hspace{2pt}+SPViT\gray{$_\text{[ECCV '22]}$}~\cite{kong2022spvit}  &        0.9  &    5  &  72.1 (\red{-0.1})    \\
  \hspace{2pt}+ToMe\gray{$_\text{[ICLR '23]}$}~\cite{bolya2022token}                 &         0.7  &  5  &    71.3 (\red{-0.9})    \\
  \hspace{2pt}+BAT\gray{$_\text{[CVPR '23]}$}~\cite{long2023beyond} &      0.8  & 5  &  72.3 (\green{+0.1})    \\
  \cmidrule(lr){1-4}
  \hspace{2pt}+MCTF$_{r=16}$  & \textbf{0.7} & 5  & \textbf{72.7 (\green{+0.5})}  \\
  \midrule
  \rowcolor{Gray}DeiT-S~\cite{touvron2021training}     & 4.6 & 22 &   79.8 (-)               \\
  \hspace{2pt}+IA-RED$^{2}$\gray{$_\text{[NeurIPS '21]}$}~\cite{pan2021ia}             & 3.2 &  22 &  79.1 (\red{-0.7})      \\
  \hspace{2pt}+DynamicViT\gray{$_\text{[NeurIPS '21]}$}~\cite{rao2021dynamicvit}              &  2.9 & 23 &  79.3 (\red{-0.5})      \\
  \hspace{2pt}+EvoViT\gray{$_\text{[AAAI '22]}$}~\cite{xu2022evo}  &  3.0  & 22  &  79.4 (\red{-0.4})    \\
  \hspace{2pt}+EViT\gray{$_\text{[ICLR '22]}$}~\cite{liang2022not}        &  3.0 & 22 &  79.5 (\red{-0.3})      \\
  \hspace{2pt}+A-ViT\gray{$_\text{[CVPR '22]}$}~\cite{yin2022vit}                     & 3.6 & 22 & 78.6 (\red{-1.2})      \\
  \hspace{2pt}+ATS\gray{$_\text{[ECCV '22]}$}~\cite{fayyaz2022adaptive}   & 2.9 & 22 &  79.7 (\red{-0.1})      \\
  \hspace{2pt}+SPViT\gray{$_\text{[ECCV '22]}$}~\cite{kong2022spvit}       & \textbf{2.6} & 22 & 79.3 (\red{-0.5})      \\
  \hspace{2pt}+ToMe\gray{$_\text{[ICLR '23]}$}~\cite{bolya2022token}                   & 2.7 &  22 &  79.4 (\red{-0.4})      \\
  \hspace{2pt}+BAT\gray{$_\text{[CVPR '23]}$}~\cite{long2023beyond} &   3.0  &  22  & 79.6 (\red{-0.2})    \\
  \cmidrule(lr){1-4}
  \hspace{2pt}+MCTF$_{r=16}$                    & \textbf{2.6} & 22 &  \textbf{80.1 (\green{+0.3})}  \\
  \bottomrule
  \end{tabular}

\vspace{-10pt}
\end{table} 

%% file: tables/table_main2.tex
\begin{table}[t!]
  \centering 
  \setlength{\tabcolsep}{3pt}
  \caption{\textbf{Comparison with other Vision Transformers}}
  \label{tab:main2}
  \vspace{-5pt}
  \begin{tabular}{lccc}
  \toprule
  \multicolumn{1}{c}{\multirow{2}{*}{Models}} & FLOPs & Params & Acc \\
  & (G) & (M) & (\%)\\
  \midrule
    PVT-Small\cite{wang2021pyramid}&	3.8&	24.5&	79.8\\
    PVT-Medium~\cite{wang2021pyramid}&	6.7&	44.2&	81.2\\
    CoaT Mini~\cite{xu2021co}&	6.8&	10.0&	80.8\\
    CoaT-Lite Small~\cite{xu2021co}&	4.0	&20.0&	81.9\\
    Swin-T~\cite{liu2021swin}&	4.5	&29.0&	81.3\\
    Swin-S~\cite{liu2021swin}&	8.7&50.0&	83.0\\
    PoolFormer-S36~\cite{yu2022metaformer}&	5.0&	31.0 &	81.4\\
    PoolFormer-M48~\cite{yu2022metaformer}&	11.6&	73.0 &	82.5\\
    \midrule
    \rowcolor{Gray}T2T-ViT${_{t}}$-14~\cite{yuan2021tokens}&	6.1&	21.5&	81.7\\
    \hspace{4pt}+MCTF$_{r=13}$&	4.2 &	21.5&	\textbf{81.8 (\green{$\boldsymbol{\uparrow}$})} \\
    \rowcolor{Gray}T2T-ViT${_{t}}$-19~\cite{yuan2021tokens}&	9.8&	39.2&	82.4\\
    \hspace{4pt}+MCTF$_{r=9}$ &	6.4 &	39.2&	\textbf{82.4} (\textbf{-}) \\
    \midrule

    \rowcolor{Gray}LV-ViT-S~\cite{jiang2021all}&	6.6&	26.2&	83.3\\
    \hspace{4pt}+EViT\gray{$_\text{[ICLR '22]}$}~\cite{liang2022not}&	4.7&	26.2& 83.0 (\red{$\boldsymbol{\downarrow}$}) \\
    \hspace{4pt}+BAT\gray{$_\text{[CVPR '23]}$}~\cite{long2023beyond} & 4.7 & 26.2& 83.1 (\red{$\boldsymbol{\downarrow}$})\\
    \hspace{4pt}+DynamicViT\gray{$_\text{[NeurIPS '21]}$}~\cite{rao2021dynamicvit}&	4.6&	26.9& 83.0 (\red{$\boldsymbol{\downarrow}$}) \\
    \hspace{4pt}+SPViT\gray{$_\text{[ECCV '22]}$}~\cite{kong2022spvit}& 4.3 & 26.2& 83.1 (\red{$\boldsymbol{\downarrow}$})  \\
    \hspace{4pt}+MCTF$_{r=12}$&	4.2&	26.2&	\textbf{83.4}  (\green{$\boldsymbol{\uparrow}$}) \\
  \bottomrule
  \end{tabular}
\vspace{-10pt}
\end{table} 

%% file: Figures/Figure6.tex
 \begin{figure*}[t]
    \centering
    \begin{subfigure}[b]{0.44\linewidth}
      \centering
       \includegraphics[trim=0 0 0 0,clip,width=\textwidth]{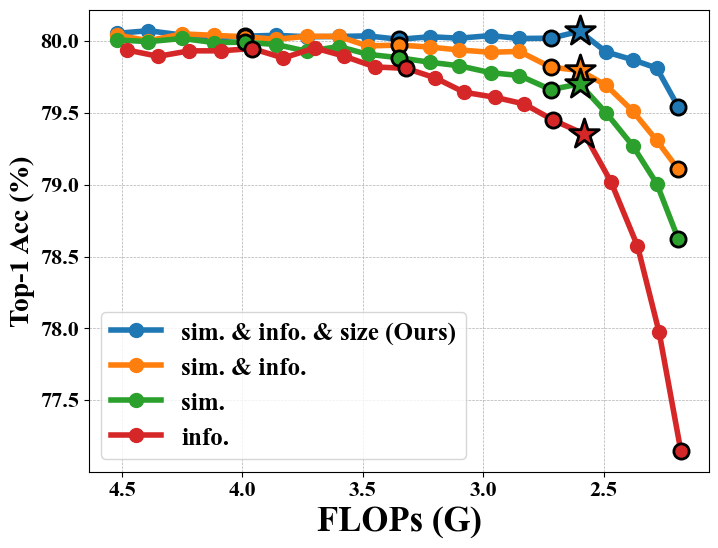}
      \caption{}
      \label{fig:abl1}
    \end{subfigure}
    \hfill
    \begin{subfigure}[b]{0.44\linewidth}
      \centering
       \includegraphics[trim=0 0 0 0,clip,width=\textwidth]{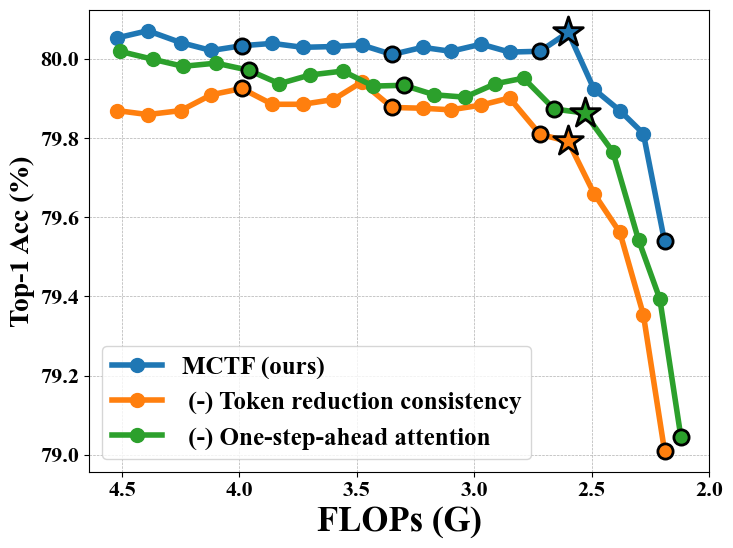}
      \caption{}
      \label{fig:abl2}
    \end{subfigure}
\vspace{-10pt}
\caption{\textbf{Ablations on (a) multi-criteria, (b) one-step-ahead-attention, and token reduction consistency.} Each marker indicates the model with $r\in[1,20]$, and we highlight $r\in\{5,10,15,20\}$ as bordered circle. We also denote the model as star when $r=16$, which is used for finetuning the model.}
\vspace{-10pt}
 \end{figure*}


%% file: tables/table_main3.tex
\begin{table}[t!]
  \centering 
  \caption{\textbf{Image classification results without training}}
  \label{tab:main3}
  \vspace{-5pt}

    \centering
    \setlength{\tabcolsep}{3pt}
    \begin{tabular}{l|cccccccc}
        \toprule
        \multicolumn{1}{c}{\multirow{2}{*}{Method}}& \multicolumn{8}{c}{$r$}\\
        \multicolumn{1}{c}{} &    Base   &   1 & 2 & 4 & 8 & 12 &16 & 20\\
        \midrule
        \rowcolor{Gray}  \multicolumn{9}{c}{\textit{DeiT-T}} \\
        ToMe \citep{bolya2022token} &  72.2 & 72.1 & 72.0 & 72.0 & 71.6 & 70.8 & 68.7 & 61.5 \\
        MCTF &  72.2 & \textbf{72.2} & \textbf{72.1} & \textbf{72.1} & \textbf{72.0} & \textbf{71.7} &\textbf{71.0} & \textbf{68.5}\\
        \midrule
        \rowcolor{Gray}  \multicolumn{9}{c}{\textit{DeiT-S}} \\
        ToMe \citep{bolya2022token} &  79.8 & 79.8 & 79.7 & 79.7 & 79.4 & 79.0 & 77.9 & 74.2\\
        MCTF & 79.8 & 79.8 & \textbf{79.8} & \textbf{79.8} & \textbf{79.8} & \textbf{79.6} &\textbf{79.2} & \textbf{78.0}\\
        \bottomrule
  \end{tabular}
\vspace{-10pt}
\end{table}

%% file: Figures/Figure7.tex
 \begin{figure*}[t!]
    \centering
      \centering
       \includegraphics[trim=0 10 0 0,clip,width=\textwidth]{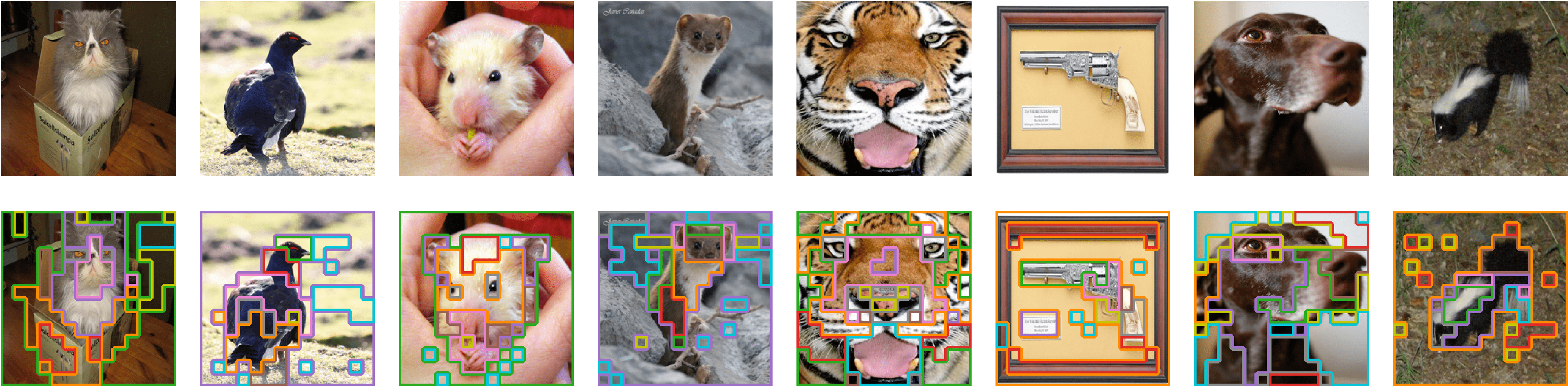}
      
\caption{\textbf{Visualization of the fused tokens with MCTF.} Given the input images of ImageNet-1K (Top), the qualitative results of MCTF with DeiT-S are provided at the bottom. The same border color of the patches indicates the fused tokens.}
\label{fig:vis}
\vspace{-10pt}
 \end{figure*}


%% file: tables/table_abl1.tex
\begin{table}[t!]
  \centering 
  \caption{\textbf{Ablations of the design choices.}}
  \label{tab:abl1}
  \vspace{-5pt}
    \centering
    \setlength{\tabcolsep}{10pt}
    \begin{tabular}{l|cc}
        \toprule
        \multicolumn{1}{c}{\multirow{2}{*}{Method}} & FLOPs $\downarrow$  & Acc $\uparrow$ \\
        \multicolumn{1}{c}{}& (G) & (\%) \\
        \midrule
        DeiT-S & 4.6 & 79.8 \\ 
        \midrule
        \rowcolor{Gray} \multicolumn{3}{c}{\textit{bipartite soft matching}}\\
        One-way & 2.6 & 80.0\\
        \rowcolor{Cyan}Bidirectional&2.6&80.1\\
        \midrule
        \rowcolor{Gray} \multicolumn{3}{c}{\textit{pooling function $\delta$}}\\
        average&2.6&80.0\\
        max&2.6&79.8\\
        \rowcolor{Cyan}weighted average&2.6&80.1\\
        \midrule
        \rowcolor{Gray} \multicolumn{3}{c}{\textit{approximation of attention map}}\\
        precise attention & 3.0 & 80.1\\
        \rowcolor{Cyan} approximated attention &2.6& 80.1\\
        \bottomrule
  \end{tabular}
\vspace{-10pt}
\end{table}

%% file: Figures/Figure8.tex
 \begin{figure}[t!]
    \centering
       \includegraphics[trim=0 0 0 0,clip,width=0.9\linewidth]{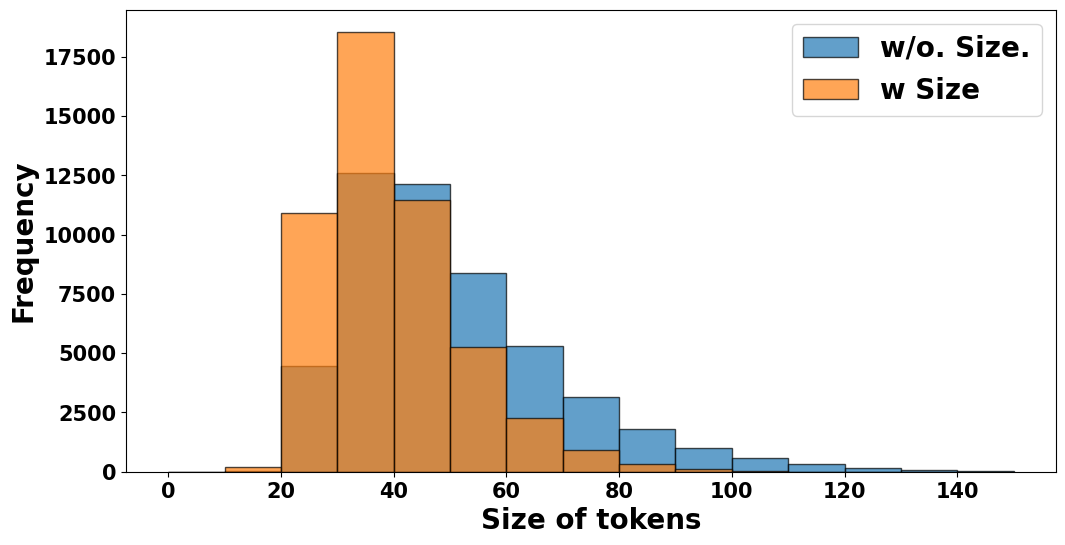}
\caption{\textbf{Histogram of the size of tokens after reduction.}}
\label{fig:hist}
\vspace{-10pt}
 \end{figure}

%% file: section/conclusion.tex
\section{Conclusion}
\label{sec:6}
In this work, we introduced the Multi-Criteria Token Fusion (MCTF), a novel strategy aimed at reducing 
the complexity inherent in ViTs while mitigating performance degradation. 
MCTF effectively discerns the relation of tokens based on multiple criteria, including similarity, informativeness, and the size of the tokens.
Our comprehensive ablation studies and detailed analyses demonstrate the efficacy of MCTF particularly with our innovative one-step-ahead attention and token reduction consistency. 
Remarkably, DeiT-T and DeiT-S with MCTF achieve considerable improvements, with +0.5\%, and +0.3\% increase in Top-1 Accuracy over the vanilla models, accompanied by 
about 44\% fewer FLOPs, respectively. 
We also observe that our MCTF outperforms all of the previous token reduction methods in diverse vision Transformers with and without training. 

\section*{Acknowledgments}
This work was supported by ICT Creative Consilience Program through the Institute of Information \& Communications Technology Planning \& Evaluation (IITP) grant funded by the Korea government (MSIT)(IITP-2024-2020-0-01819), the National Research Foundation of Korea (NRF) grant funded by the Korea government (MSIT)(NRF-2023R1A2C2005373), and a grant of the Korea Health Technology R\&D Project through the Korea Health Industry Development Institute (KHIDI) funded by the Ministry of Health \& Welfare Republic of Korea (HR20C0021).

%% file: tables/table_sup5.tex
\begin{table}[h]
  \centering 
  \caption{\textbf{Sensitivity analysis on the hyper-parameters.}}
  \label{tab:sup5} 
  \setlength{\tabcolsep}{4pt}
  \begin{tabular}{c|cccccc}
  \toprule
  $\tau_\text{sim}$  &   1  & 1/5 & 1/10 & 1/20 & 1/40 & 1/100 \\
  \midrule
  acc.             & \cellcolor{Cyan}80.1 & 79.6 & 79.2 & 78.6 & 78.1 & 77.5  \\
  \bottomrule
  \multicolumn{1}{c}{}\\
  \toprule
  $\tau_\text{info}$  &   1  & 1/5 & 1/10 & 1/20 & 1/40 & 1/100 \\
  \midrule
  acc.             & 78.7 & 79.8 & 80.0 & \cellcolor{Cyan}80.1 & 80.0 & 79.8  \\
  \bottomrule
  \multicolumn{1}{c}{}\\
  \toprule
  $\tau_\text{size}$  & 1  & 1/5 & 1/10 & 1/20 & 1/40 & 1/100 \\
  \midrule
  acc.            & 79.5 & 79.8 & 80.0 & 80.0 & \cellcolor{Cyan}80.1 & 80.0\\
  \bottomrule
  
  \end{tabular}
\end{table}

%% file: tables/table_sup4.tex
\begin{table*}[t]
  \centering 
  \caption{\textbf{Cosine similarity between the class tokens with and without MCTF per block.}}
  \label{tab:sup4} 
  \setlength{\tabcolsep}{3pt}
  \begin{tabular}{ccc|cccccccccccc}
  \toprule
   \multicolumn{3}{c|}{Criteria}&\multicolumn{12}{c}{Block index}\\
  $\mathbf{W}^\text{sim}$ & $\mathbf{W}^\text{info}$ & $\mathbf{W}^\text{size}$& 1 & 2 & 3 & 4 & 5 & 6 & 7& 8 & 9 & 10 & 11 & 12\\
     \midrule
     $\checkmark$ &&&1.0000 & 1.0000 & 1.0000 & 1.0000 & 0.9999 & 0.9996 & 0.9988 & 0.9973 & 0.9933 & 0.9870 & 0.9837 & 0.9695 \\
     &$\checkmark$& &  1.0000 & 1.0000 & 0.9999 & 0.9996 & 0.9992 & 0.9976 & 0.9939 & 0.9887 & 0.9750 & 0.9550 & 0.9470 & 0.9153 \\
     &&$\checkmark$ &1.0000 & 1.0000 & 0.9998 & 0.9996 & 0.9991 & 0.9968 & 0.9913 & 0.9812 & 0.9575 & 0.9141 & 0.9040 & 0.8546\\
     $\checkmark$ &$\checkmark$& & 1.0000 & 1.0000 & 1.0000 & 1.0000 & 0.9999 & 0.9997 & 0.9992 & 0.9982 & 0.9958 & 0.9925 & 0.9907 & 0.9833\\
     $\checkmark$ &$\checkmark$&$\checkmark$ & 1.0000 & 1.0000 & 1.0000 & 1.0000 & 0.9999 & \textbf{0.9997} & \textbf{0.9992} & \textbf{0.9984} & \textbf{0.9961} & \textbf{0.9929} & \textbf{0.9914} & \textbf{0.9844}\\
  \bottomrule
  \end{tabular}
\end{table*} 

%% file: Figures/onestepahead_qa.tex
 \begin{figure*}[h]
    \centering
      \centering
       \includegraphics[trim=0 0 0 0,clip,width=\textwidth]{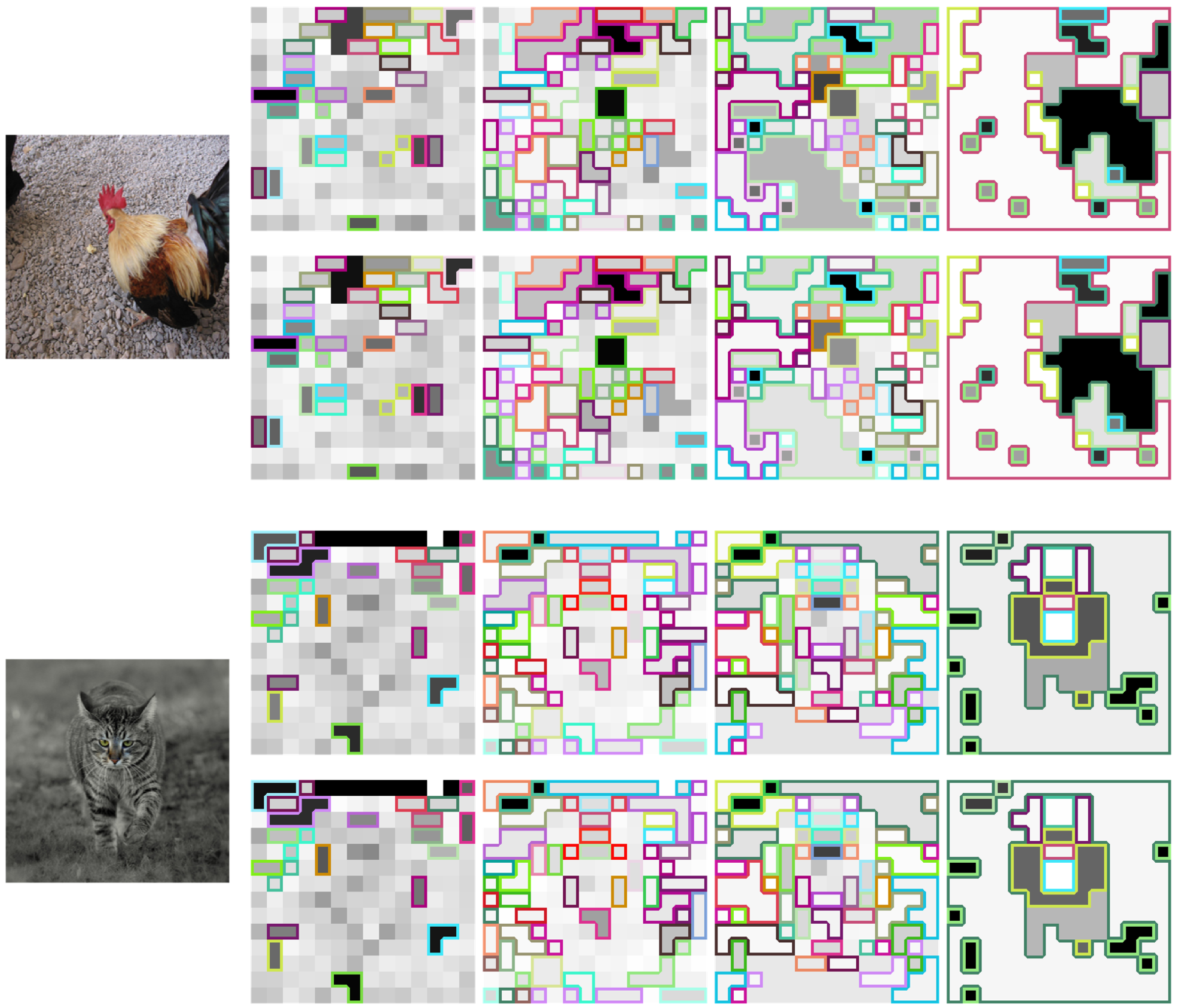}
\caption{\textbf{Comparison of approximated and precise attention map for $\hat{\mathbf{A}}^{l+1}$.} Given the left image, we visualize the (Top) approximated attention map and (Bottom) precise attention map.}
\label{fig:onestepahead_qa}
\vspace{0pt}
 \end{figure*}


%% file: tables/table_sup1.tex
\begin{table}[h]
  \centering 
  \caption{\textbf{Detailed results of MCTF with DeiT after finetuning with $r=16$.}}
  \label{tab:sup1} 
  \setlength{\tabcolsep}{8pt}
  \begin{tabular}{c|cccc|cccc}
  \toprule
  \multicolumn{1}{c|}{\multirow{3}{*}{$r$}} &\multicolumn{4}{c|}{DeiT-T} & \multicolumn{4}{c}{DeiT-S}\\
  
  & \multicolumn{2}{c}{FLOPs} & \multicolumn{2}{c|}{Top-1 Acc}  & \multicolumn{2}{c}{FLOPs} & \multicolumn{2}{c}{Top-1 Acc} \\
  & (G) & $\downarrow$ (\%) & (\%) & $\Delta$ & (G) & $\downarrow$ (\%)& (\%) & $\Delta$\\
 
  \midrule
     \rowcolor{Gray} Base & 1.26 & - & 72.2 & - & 4.61 & - & 79.8 & - \\
     \midrule
     1  &  1.24  & 1.59 &  72.92  & +0.72  &  4.52 & 1.95 &  80.06 & +0.26\\
     2  &  1.20  & 4.76 &  72.91  & +0.71  &  4.39 & 4.77 &  80.07 & +0.27\\
     3  &  1.17  & 7.14 &  72.92  & +0.72  &  4.25 & 7.81 &  80.04 & +0.24\\
     4  &  1.13  & 10.32 &  72.91  & +0.71  &  4.12 & 10.63 &  80.02 & +0.22\\
     5  &  1.09  & 13.49 &  72.92  & +0.72  &  3.99 & 13.45 &  80.03 & +0.23\\
     6  &  1.06  & 15.87 &  72.92  & +0.72  &  3.86 & 16.27 &  80.04 & +0.24\\
     7  &  1.02  & 19.05 &  72.91  & +0.71  &  3.73 & 19.09 &  80.03 & +0.23\\
     8  &  0.98  & 22.22 &  72.94  & +0.74  &  3.60 & 21.91 &  80.03 & +0.23\\
     9  &  0.95  & 24.60 &  72.86  & +0.66  &  3.48 & 24.51 &  80.04 & +0.24\\
    10  &  0.91  & 27.78 & 72.77  & +0.57  &  3.35 & 27.33 &  80.01 & +0.21\\
    11  &  0.88  & 30.16 &  72.81 & +0.61  &  3.22 & 30.15 &  80.03 & +0.23\\
    12  &  0.84  & 33.33 &  72.76 & +0.56  &  3.10 & 32.75 &  80.02 & +0.22\\
    13  &  0.81  & 35.71 &  72.73  & +0.53  &  2.97 & 35.57 &  80.04 & +0.24\\
    14  &  0.78  & 38.10 &  72.71  & +0.51  &  2.85 & 38.18 &  80.02 & +0.22\\
    15  &  0.74  & 41.27 &  72.72  & +0.52  &  2.72 & 41.00 &  80.02 & +0.22\\
    16  &  \cellcolor{Cyan}0.71  & \cellcolor{Cyan}43.65 &  \cellcolor{Cyan}72.66  & \cellcolor{Cyan}+0.46  & \cellcolor{Cyan}2.60 & \cellcolor{Cyan}43.60 & \cellcolor{Cyan}80.07 & \cellcolor{Cyan}+0.27\\
    17  &  0.68  & 46.03 &  72.38  & +0.18  &  2.49 & 45.99 &  79.93 & +0.13\\
    18  &  0.65  & 48.41 &  72.07  & -0.13  &  2.38 & 48.37 &  79.87 & +0.07\\
    19  &  0.62  & 50.79 &  71.86  & -0.34  &  2.28 & 50.54 &  79.81 & +0.01\\
    20  &  0.60  & 52.38 &  71.35  & -0.85  &  2.19 & 52.49 &  79.54 & -0.26\\

  \bottomrule
  \end{tabular}
\end{table} 

%% file: tables/table_sup2.tex
\begin{table}[h]
  \centering 
  \caption{\textbf{Detailed results of MCTF with DeiT without any addtional training.}}
  \label{tab:sup2} 
  \setlength{\tabcolsep}{8pt}
  \begin{tabular}{c|cccc|cccc}
  \toprule
  \multicolumn{1}{c|}{\multirow{3}{*}{$r$}} &\multicolumn{4}{c|}{DeiT-T} & \multicolumn{4}{c}{DeiT-S}\\
  
 & \multicolumn{2}{c}{FLOPs} & \multicolumn{2}{c|}{Top-1 Acc}  & \multicolumn{2}{c}{FLOPs} & \multicolumn{2}{c}{Top-1 Acc} \\
  & (G) & $\downarrow$ (\%) & (\%) & $\Delta$ & (G) & $\downarrow$ (\%)& (\%) & $\Delta$\\
 
  \midrule
     \rowcolor{Gray} Base & 1.26 & - & 72.2 & -  & 4.61 & - & 79.8 & - \\
     \midrule
     1  &  1.24  & 1.59 &  72.15  & -0.05  &  4.52 & 1.95 &  79.78 & -0.02\\
     2  &  1.20  & 4.76 &  72.09  & -0.11  &  4.39 & 4.77 &  79.81 & +0.01\\
     3  &  1.17  & 7.14 &  72.06  & -0.14  &  4.25 & 7.81 &  79.79 & -0.01\\
     4  &  1.13  & 10.32 &  72.06  & -0.14  &  4.12 & 10.63 &  79.83 & +0.03\\
     5  &  1.09  & 13.49 &  72.06  & -0.14  &  3.99 & 13.45 &  79.81 & +0.01\\
     6  &  1.06  & 15.87 &  72.00  & -0.20  &  3.86 & 16.27 &  79.74 & -0.06\\
     7  &  1.02  & 19.05 &  72.00  & -0.20  &  3.73 & 19.09 &  79.72 & -0.08\\
     8  &  0.98  & 22.22 &  71.98  & -0.22  &  3.60 & 21.91 &  79.76 & -0.04\\
     9  &  0.95  & 24.60 &  71.92  & -0.28  &  3.48 & 24.51 &  79.68 & -0.12\\
    10  &  0.91  & 27.78 &  71.88  & -0.32  &  3.35 & 27.33 &  79.64 & -0.16\\
    11  &  0.88  & 30.16 &  71.82  & -0.38  &  3.22 & 30.15 &  79.61 & -0.19\\
    12  &  0.84  & 33.33 &  71.72  & -0.48  &  3.10 & 32.75 &  79.62 & -0.18\\
    13  &  0.81  & 35.71 &  71.61  & -0.59  &  2.97 & 35.57 &  79.54 & -0.26\\
    14  &  0.78  & 38.10 &  71.50  & -0.70  &  2.85 & 38.18 &  79.41 & -0.39\\
    15  &  0.74  & 41.27 &  71.28  & -0.92  &  2.72 & 41.00 &  79.36 & -0.44\\
    16  &  0.71  & 43.65 &  70.99  & -1.21  &  2.60 & 43.60 &  79.21 & -0.59\\
    17  &  0.68  & 46.03 &  70.62  & -1.58  &  2.49 & 45.99 &  79.06 & -0.74\\
    18  &  0.65  & 48.41 &  70.01  & -2.19  &  2.38 & 48.37 &  78.80 & -1.00\\
    19  &  0.62  & 50.79 &  69.41  & -2.79  &  2.28 & 50.54 &  78.63 & -1.17\\
    20  &  0.60  & 52.38 &  68.52  & -3.68  &  2.19 & 52.49 &  78.06 & -1.74\\

  \bottomrule
  \end{tabular}
\end{table} 

%% file: tables/table_sup3.tex
\begin{table}[h]
  \centering 
  \caption{\textbf{Detailed results of MCTF with T2T-ViT and LV-ViT.}}
  \label{tab:sup3} 
  \begin{tabular}{c|cccc|cccc|cccc}
  \toprule
  \multicolumn{1}{c|}{\multirow{3}{*}{$r$}} &\multicolumn{4}{c|}{T2T-ViT$_t$-14} & \multicolumn{4}{c|}{T2T-ViT$_t$-19}& \multicolumn{4}{c}{LV-ViT-S}\\
  
 & \multicolumn{2}{c}{FLOPs} & \multicolumn{2}{c|}{Top-1 Acc}  & \multicolumn{2}{c}{FLOPs} & \multicolumn{2}{c|}{Top-1 Acc} & \multicolumn{2}{c}{FLOPs} & \multicolumn{2}{c}{Top-1 Acc} \\
  & (G) & $\downarrow$ (\%) & (\%) & $\Delta$ & (G) & $\downarrow$ (\%)& (\%) & $\Delta$ & (G) & $\downarrow$ (\%)& (\%) & $\Delta$\\
 
  \midrule
       \rowcolor{Gray} Base & 6.11 & - & 81.7 & - &  9.81 & - & 82.4 & - & 6.50 & - & 83.3 & -\\
     \midrule
     1  &  6.00  & 1.80 & 81.84 & +0.14 & 9.50 & 3.16 & 82.42 & +0.02 & 6.34 & 2.46 & 83.51 & +0.21 \\
     2  &  5.84  & 4.42 & 81.85 & +0.15 & 9.10 & 7.24 & 82.43 & +0.03 & 6.14 & 5.54 & 83.53 & +0.23 \\
     3  &  5.69  & 6.87 & 81.82 & +0.12 & 8.71 & 11.21 & 82.40 & $\pm$0.00 & 5.93 & 8.87 & 83.50 & +0.20 \\
     4  &  5.53  & 9.49 & 81.83 & +0.13 & 8.32 & 15.19 & 82.43 & +0.03 & 5.73 & 11.85 & 83.51 & +0.21 \\
     5  &  5.38  & 11.95 & 81.83 & +0.13 & 7.94 & 19.06 & 82.39 & -0.01 & 5.52 & 15.08 & 83.48 & +0.18 \\
     6  &  5.23  & 14.40 & 81.84 & +0.14 & 7.56 & 22.94 & 82.43 & +0.03 & 5.32 & 18.15 & 83.48 & +0.18 \\
     7  &  5.07  & 17.02 & 81.84 & +0.14 & 7.18 & 26.81 & 82.41 & +0.01 & 5.12 & 21.23 & 83.52 & +0.22 \\
     8  &  4.92  & 19.48 & 81.80 & +0.10 & 6.81 & 30.58 & 82.42 & +0.02 & 4.93 & 24.15 & 83.47 & +0.17 \\
     9  &  4.78  & 21.77 & 81.81 & +0.11 & \cellcolor{Cyan}6.44 & \cellcolor{Cyan}34.35 & \cellcolor{Cyan}82.39 & \cellcolor{Cyan}-0.01 & 4.73 & 27.23 & 83.48 & +0.18 \\
    10  &  4.63  & 24.22 & 81.76 & +0.06 & 6.08 & 38.02 & 82.27 & -0.13 & 4.54 & 30.15 & 83.47 & +0.17 \\
    11  &  4.48  & 26.68 & 81.81 & +0.11 & 5.74 & 41.49 & 82.25 & -0.15 & 4.35 & 33.08 & 83.44 & +0.14 \\
    12  &  4.34  & 28.97 & 81.80 & +0.10 & 5.45 & 44.44 & 82.02 & -0.38 & \cellcolor{Cyan}4.16 & \cellcolor{Cyan}36.00 & \cellcolor{Cyan}83.37 & \cellcolor{Cyan}+0.07\\
    13  &  \cellcolor{Cyan}4.19  & \cellcolor{Cyan}31.42 & \cellcolor{Cyan}81.76 & \cellcolor{Cyan}+0.06& 5.21 & 46.89 & 81.86 & -0.54 & 3.98 & 38.77 & 83.23 & -0.07 \\
    14  &  4.05  & 33.72 & 81.69 & -0.01 & 5.00 & 49.03 & 81.38 & -1.02 & 3.83 & 41.08 & 83.03 & -0.27 \\
    15  &  3.92  & 35.84 & 81.51 & -0.19 & 4.82 & 50.87 & 80.85 & -1.55 & 3.69 & 43.23 & 82.72 & -0.58 \\
    16  &  3.80  & 37.81 & 81.48 & -0.22 & 4.67 & 52.40 & 80.46 & -1.94 & 3.58 & 44.92 & 82.28 & -1.02 \\
    17  &  3.70  & 39.44 & 81.22 & -0.48 & 4.53 & 53.82 & 80.29 & -2.11 & 3.48 & 46.46 & 81.81 & -1.49 \\
    18  &  3.61  & 40.92 & 80.93 & -0.77 & 4.41 & 55.05 & 79.58 & -2.82 & 3.38 & 48.00 & 81.01 & -2.29 \\
    19  &  3.53  & 42.23 & 80.67 & -1.03 & 4.30 & 56.17 & 79.29 & -3.11 & 3.31 & 49.08 & 80.73 & -2.57 \\
    20  &  3.45  & 43.54 & 80.11 & -1.59 & 4.20 & 57.19 & 78.41 & -3.99 & 3.23 & 50.31 & 79.85 & -3.45 \\

  \bottomrule
  \end{tabular}
\end{table} 

%% file: supple/implementation.tex
\section{Implementation details}
\label{sec:a}
For comparison with previous works, we evaluate MCTF with DeiT~\citep{touvron2021training} on ImageNet-1K\footnote{\copyright 2020 Stanford Vision Lab, Stanford University, Princeton University}~\citep{deng2009imagenet} in three settings; (1) finetuning, (2) training from scratch, and (3) without training.
First, following~\citep{rao2021dynamicvit, kong2022spvit,yin2022vit}, we finetune the model with the official pre-trained weights for 30 epochs. Note that we opt for the least epochs among previous works (30 for DynamicViT~\citep{rao2021dynamicvit}, 60 for SPViT~\citep{kong2022spvit}, 100 for A-ViT~\citep{yin2022vit}), but achieves the best performance. For finetuning, the learning rate is initially set to 5e-5 and decreases to 1e-6 by the cosine annealing~\citep{loshchilov2016sgdr} without any warmup or cooldown process.
Next, to compare MCTF with the methods~\citep{liang2022not, bolya2022token} trained from scratch, we train the model equipped with MCTF following the default configuration used in DeiT~\citep{touvron2021training}. Lastly, we simply apply MCTF on the pre-trained model without any finetuning and training. Since there are no model updates without training, self-distillation is not used in this case. \\
For measuring the efficiency, we use \texttt{fvcore}\footnote{\copyright 2019 Facebook, Inc, Licensed under the Apache License 2.0
} to report FLOPs under one RTX3090, and, regarding hyper-parameters for MCTF, we use $\tau_\mathbf{a} = 1/25, \tau_\mathbf{s} = 1/10$ in all experiments. Also, to finetune the model, we opt for ($r =16, \alpha = 1$) for DeiT-T, DeiT-S, and ($r=14, \alpha = 2$) for Deit-B. 

%% file: supple/experiments.tex
\section{More experiments}
\label{sec:b}
\subsection{Detailed results of image classification.}
We provide more detailed results with varying $r$ after finetuning the model with MCTF ($r=16$). The results are summarized in~\Cref{tab:sup1}. As discussed in the main paper, the model trained with $r=16$ shows competitive or even better performance across various $r$.
\input{supple/table_s1}
\subsection{Image classification results without training.}
Detailed results with MCTF in DeiT~\citep{touvron2021training} without any finetuning and training are presented in~\Cref{tab:sup2}. 
\input{supple/table_s2}
\subsection{MCTF with other ViTs.}
To show the applicability of MCTF on other ViTs including more huge models, we evaluate MCTF with pre-trained ViTs~\citep{dosovitskiy2020image} in~\citep{steiner2021train} known as AugReg. The results are summarized in~\Cref{tab:sup4}. Since ViT-L consists of 24 blocks, the maximum number of reduced tokens per block $r$ is 8 with ViT-L. 
\input{supple/table_s3}


%% file: supple/table_s1.tex
\begin{table}[h]
  \centering 
  \caption{\textbf{Detailed results after finetuning with $r=16$.}}
  \label{tab:sup1} 
  \setlength{\tabcolsep}{8pt}
  \begin{tabular}{c|ccc|ccc|ccc}
  \toprule
  \multicolumn{1}{c|}{\multirow{3}{*}{$r$}}  &\multicolumn{3}{c|}{DeiT-T}& \multicolumn{3}{c|}{DeiT-S}& \multicolumn{3}{c}{DeiT-B}\\
\multicolumn{1}{c|}{} & FLOPs & thr. & acc & FLOPs & thr. & acc.  & FLOPs & thr. & acc.\\
  \midrule
     \rowcolor{Gray} Base & 1.26 & 3181 & 72.2 & 4.61 & 1262 & 79.8 & 17.58 & 412 & 81.8\\
     \midrule

     1  &  1.24  &  2829  &  73.0  & 4.51  &  1143  &  79.9  &  17.16  &  389  &  82.0 \\
     2  &  1.20  &  2904  &  73.0  & 4.38  &  1174  &  79.9  &  16.65  &  401  &  82.0 \\
     3  &  1.16  &  2995  &  73.0  & 4.24  &  1211  &  79.9  &  16.14  &  415  &  82.0 \\
     4  &  1.12  &  3084  &  73.0  & 4.10  &  1249  &  80.0  &  15.64  &  427  &  82.0 \\
     5  &  1.08  &  3188  &  73.0  & 3.97  &  1289  &  80.0  &  15.13  &  442  &  82.0 \\
     6  &  1.05  &  3296  &  72.9  & 3.83  &  1333  &  79.9  &  14.63  &  457  &  81.9 \\
     7  &  1.01  &  3441  &  72.9  & 3.70  &  1391  &  79.9  &  14.13  &  475  &  81.8 \\
     8  &  0.97  &  3579  &  72.8  & 3.56  &  1445  &  79.8  &  13.63  &  494  &  81.9 \\
     9  &  0.93  &  3719  &  72.8  & 3.43  &  1507  &  79.9  &  13.13  &  510  &  81.9 \\
    10  &  0.90  &  3879  &  72.8  & 3.30  &  1572  &  79.8  &  12.63  &  534  &  81.8 \\
    11  &  0.86  &  4022  &  72.8  & 3.17  &  1635  &  79.8  &  12.14  &  557  &  81.8 \\
    12  &  0.83  &  4200  &  72.7  & 3.04  &  1704  &  79.9  &  11.65  &  578  &  81.8 \\
    13  &  0.79  &  4376  &  72.8  & 2.91  &  1777  &  79.8  &  11.16  &  605  &  81.7 \\
    14  &  0.76  &  4565  &  72.8  & 2.79  &  1858  &  79.8  &  10.67  &  646  &  81.7 \\
    15  &  0.72  &  4775  &  72.7  & 2.66  &  1940  &  79.8  &  10.18  &  663  &  81.5 \\
    16  &  0.69  &  4992  &  72.6  & 2.53  &  2016  &  79.9  &   9.60  &  695  &  81.3 \\

  \bottomrule
  \end{tabular}
\end{table} 

%% file: supple/table_s2.tex
\begin{table}[h]
  \centering 
  \caption{\textbf{Detailed results without training.}}
  \label{tab:sup2} 
  \setlength{\tabcolsep}{8pt}
  \begin{tabular}{c|ccc|ccc|ccc}
  \toprule
  \multicolumn{1}{c|}{\multirow{3}{*}{$r$}}  &\multicolumn{3}{c|}{DeiT-T}& \multicolumn{3}{c|}{DeiT-S}& \multicolumn{3}{c}{DeiT-B}\\
\multicolumn{1}{c|}{} & FLOPs & thr. & acc & FLOPs & thr. & acc.  & FLOPs & thr. & acc.\\
  \midrule
     \rowcolor{Gray}Base  & 1.26 & 3181 & 72.2 & 4.61 & 1262 & 79.8  & 17.58 & 412 & 81.8  \\
     1  &  1.24  &  2795  &  72.1  &  4.51  &  1153  &  79.8  &  17.16  &  386  &  81.8 \\
     2  &  1.20  &  2868  &  72.1  &  4.38  &  1186  &  79.8  &  16.65  &  398  &  81.7 \\
     3  &  1.16  &  2972  &  72.1  &  4.24  &  1224  &  79.8  &  16.14  &  412  &  81.7 \\
     4  &  1.12  &  3061  &  72.1  &  4.10  &  1263  &  79.8  &  15.64  &  424  &  81.8 \\
     5  &  1.08  &  3168  &  72.1  &  3.97  &  1305  &  79.7  &  15.13  &  438  &  81.8 \\
     6  &  1.05  &  3275  &  72.0  &  3.83  &  1347  &  79.7  &  14.63  &  453  &  81.6 \\
     7  &  1.01  &  3418  &  71.9  &  3.70  &  1406  &  79.7  &  14.13  &  471  &  81.5 \\
     8  &  0.97  &  3545  &  71.9  &  3.56  &  1458  &  79.7  &  13.63  &  489  &  81.5 \\
     9  &  0.93  &  3699  &  71.8  &  3.43  &  1516  &  79.6  &  13.13  &  507  &  81.3 \\
    10  &  0.90   &  3842  &  71.7  &  3.30  &  1583  &  79.6  &  12.63  &  530  &  81.3 \\
    11  &  0.86  &  3995  &  71.7  &  3.17  &  1646  &  79.5  &  12.14  &  552  &  81.2 \\
    12  &  0.83  &  4160  &  71.7  &  3.04  &  1720  &  79.4  &  11.65  &  574  &  81.0 \\
    13  &  0.79  &  4346  &  71.5  &  2.91  &  1792  &  79.5  &  11.16  &  600  &  80.8 \\
    14  &  0.76  &  4527  &  71.3  &  2.79  &  1875  &  79.3  &  10.67  &  629  &  80.7 \\
    15  &  0.72  &  4749  &  70.9  &  2.66  &  1956  &  79.1  &  10.18  &  658  &  80.3 \\
    16  &  0.69  &  4992  &  70.5  &  2.53  &  2016  &  79.1  &   9.69  &  690  &  79.9 \\

  \bottomrule
  \end{tabular}
\end{table} 

%% file: supple/table_s3.tex
\begin{table}[h]
  \centering 
  \caption{\textbf{MCTF with other pretrained ViTs.}}
  \label{tab:sup3} 
  \setlength{\tabcolsep}{8pt}
  \begin{tabular}{c|cccc|cccc}
  
  \toprule
 $r$ & Models & FLOPs & thr. & acc. & Models & FLOPs & thr. & acc. \\
  \midrule
     \rowcolor{gray} - & ViT-Ti & 1.26 & 3386 & 75.5 & ViT-S  &  4.61  &  1265  &  81.4 \\
     1  & ViT-Ti &  1.24 &  2884  &  75.4  & ViT-S & 4.51 &  1154  &  81.4 \\
     2  & ViT-Ti &  1.20 &  2964  &  75.4  & ViT-S & 4.38 &  1187  &  81.4 \\
     4  & ViT-Ti &  1.12 &  3151  &  75.3  & ViT-S & 4.10 &  1263  &  81.3 \\
     8  & ViT-Ti &  0.97 &  3651  &  75.2  & ViT-S & 3.56 &  1460  &  81.0 \\
    12  & ViT-Ti &  0.83 &  4288  &  74.7  & ViT-S & 3.04 &  1711  &  80.5 \\
    16  & ViT-Ti &  0.69 &  5073  &  73.3  & ViT-S & 2.53 &  2039  &  79.2 \\

     \multicolumn{1}{c}{}\\
  \toprule
 $r$ & Models & FLOPs & thr. & acc. & Models & FLOPs & thr. & acc. \\
  \midrule
     \rowcolor{gray} - & ViT-B & 17.58 & 410 & 84.5 & ViT-L & 61.60 & 125 & 85.7 \\
      1  & ViT-B & 17.16  &  393  &  84.5  & ViT-L &  58.13  &  124  &  85.7  \\
      2  & ViT-B & 16.65  &  405  &  84.5  & ViT-L &  54.43  &  134  &  85.7  \\
      4  & ViT-B & 15.64  &  431  &  84.4  & ViT-L &  47.10  &  154  &  85.6  \\
      8  & ViT-B & 13.63  &  497  &  84.1  & ViT-L &  32.67  &  223  &  84.7  \\
     12  & ViT-B & 11.65  &  582  &  83.6  & \multicolumn{4}{c}{-}  \\
     16  & ViT-B &  9.69  &  698  &  82.3  & \multicolumn{4}{c}{-} \\

  \multicolumn{1}{c}{}\\

  \multicolumn{1}{c}{}\\

  \end{tabular}
\end{table} 

%% file: supple/analyses.tex
\section{More analyses}
\label{sec:c}
\subsection{Sensitivity analysis on hype-rparameters.}
To analyze the sensitivity of MCTF under varying hyper-parameters, we compare the accuracy according to the temperature parameter $\tau_\mathbf{a}, \tau_\mathbf{s}$, and the weight parameter  $\alpha$ for self-distillation. While evaluating each parameter, other hyper-parameters are set to default values mentioned in~\Cref{sec:a}, including $r=16$. Same as analyses in the main paper, we run the experiments with DeiT-S. 
\input{supple/table_s4}


\subsection{Qualitative results of MCTF.}
In this subsection, we present the qualitative results of MCTF in~\Cref{fig:vis1} and~\Cref{fig:vis2}. We visualize the fused tokens after the last block. 
Since the tokens are merged with multi-criteria (\textit{e.g.}, similarity, informativeness, size), we maintain the more diverse informative tokens in the foreground object. 
For instance, compared to the tokens in the background, the tokens in the foreground objects are less fused with the moderate size lessening the information loss.

\input{supple/figure_s1}

%% file: supple/table_s4.tex
\begin{table}[h]
  \centering 
  \caption{\textbf{Sensitivity analysis on the hyper-parameters.}}
  \label{tab:sup4} 
  \setlength{\tabcolsep}{8pt}
  \begin{tabular}{c|ccccccc}
  
  \toprule
  $\tau_\mathbf{a}$  &   1  &  1/2 &  1/5 & 1/10 &       1/25             & 1/100 \\
  \midrule
  acc.             & 79.1 & 79.4 & 79.7 & 79.7 &  \cellcolor{gray} 79.9 & 79.8  \\
  \bottomrule
  \multicolumn{1}{c}{}\\
  \toprule
  $\tau_\mathbf{s}$  &   1  & 1/2  & 1/5 & 1/10 & 1/25 & 1/100 \\
  \midrule
  acc.            & 79.6 & 79.7 & 79.8 & \cellcolor{gray} 79.9 & 79.8 & 79.8 \\
  \bottomrule
  \multicolumn{1}{c}{}\\
  \toprule
  $\alpha$           &  0.1 &  0.5 & 1 & 2 & 5 & 10 \\
  \midrule
  acc.          & 79.6 &  79.7 & \cellcolor{gray} 79.9 & 79.8 & 79.7 & 79.5 &  \\
  \bottomrule
  
  \end{tabular}
\end{table} 

%% file: supple/figure_s1.tex


 \begin{figure}[t!]
    \centering
       \includegraphics[trim=0 0 0 0,clip,height=0.95\textheight]{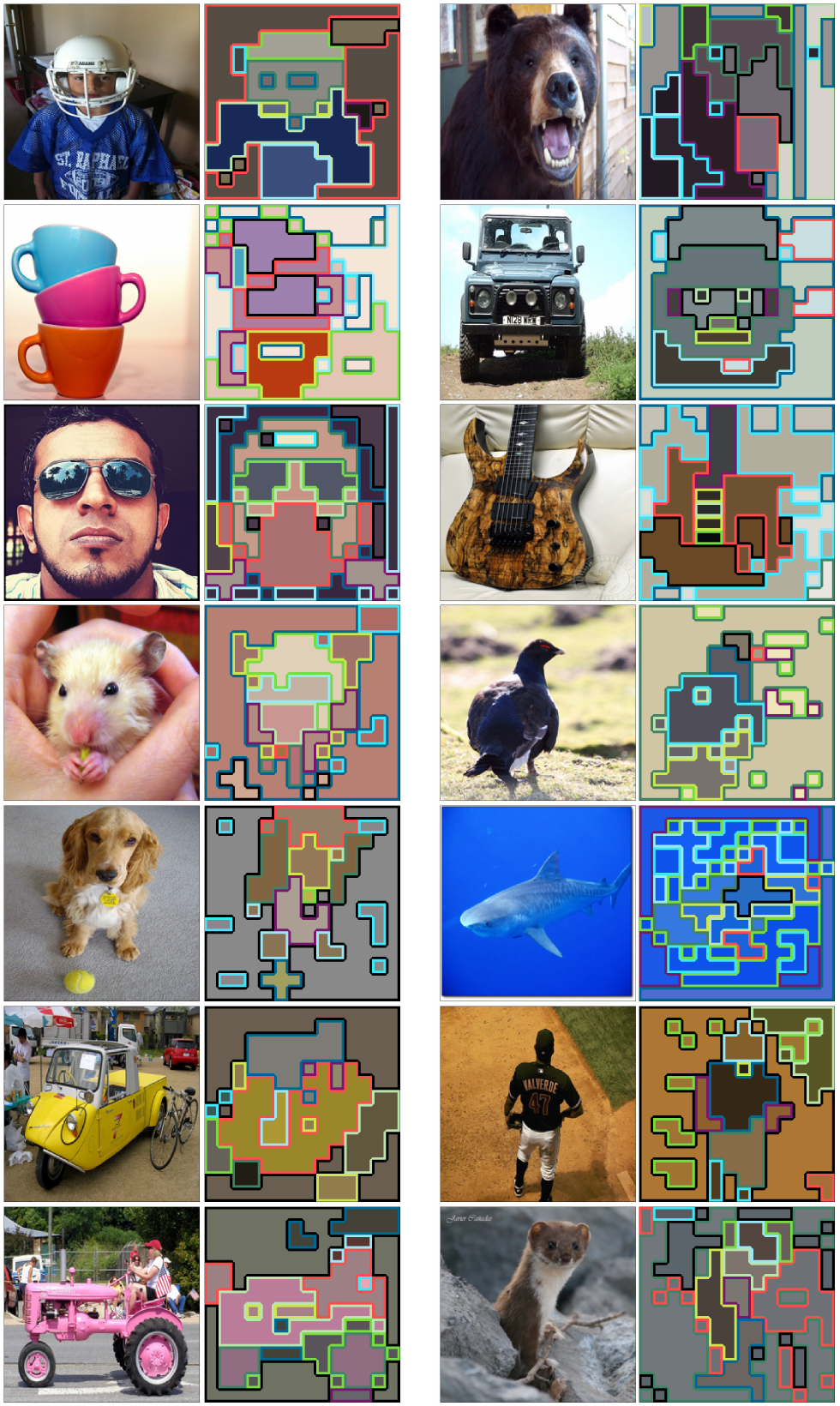}
      \caption{\textbf{Visualization of fused tokens.}}
      \label{fig:vis1}
 \end{figure}

 \begin{figure}[t!]
    \centering
       \includegraphics[trim=0 0 0 0,clip,height=0.95\textheight]{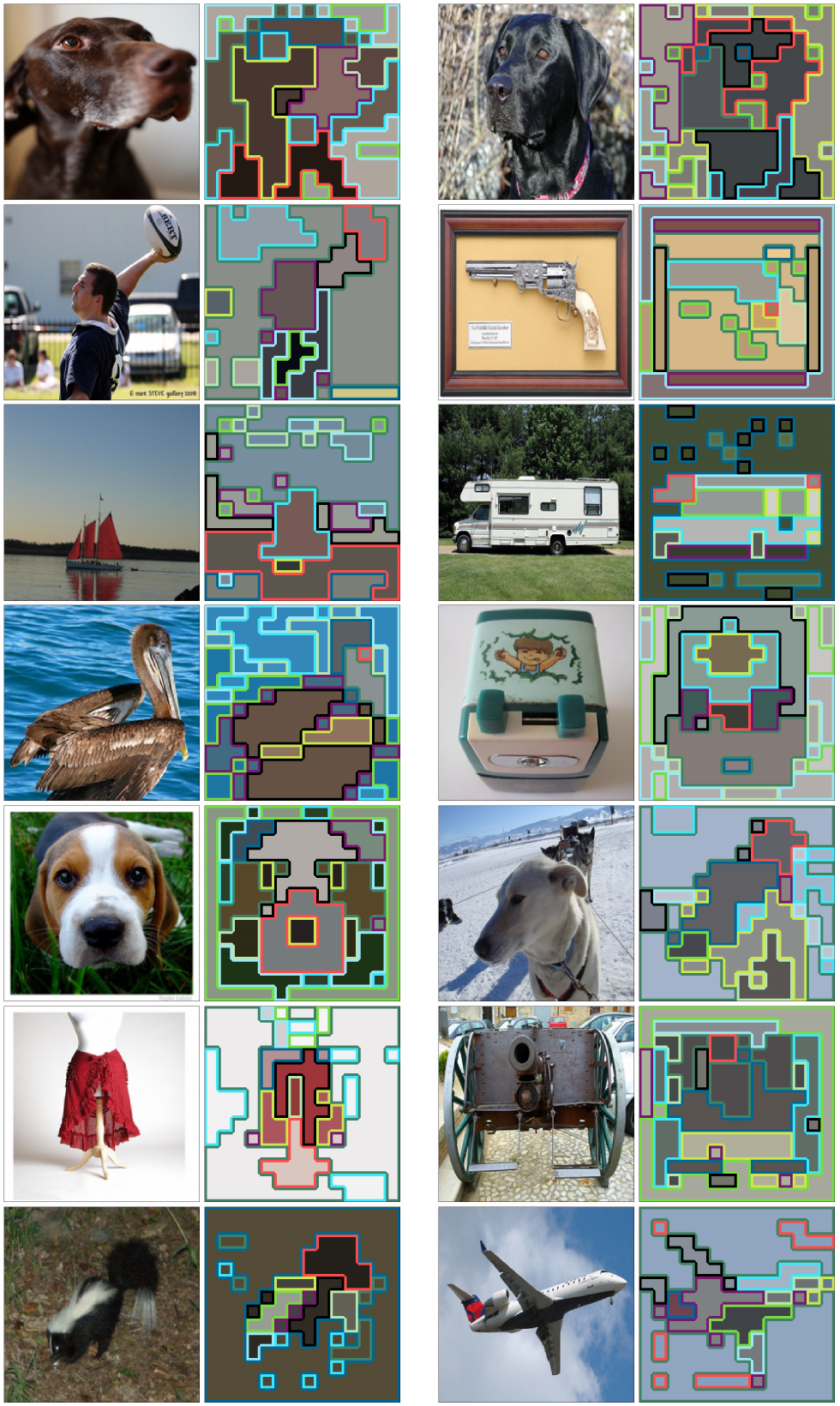}
      \caption{\textbf{Visualization of fused tokens.}}
      \label{fig:vis2}
 \end{figure}